\documentclass[journal]{IEEEtran}

\IEEEoverridecommandlockouts
\usepackage[english]{babel}
\usepackage{microtype}
\usepackage{graphicx}
\usepackage{booktabs}
\usepackage{bm}
\usepackage{tensor}
\usepackage{amsfonts}
\usepackage{url}
\usepackage{amssymb}
\usepackage{amsmath}
\usepackage{mathtools}
\usepackage{breqn}
\usepackage{siunitx}
\usepackage{xcolor}
\usepackage{afterpage}
\usepackage{soul}
\usepackage{tabularx}
\usepackage{subfigure}
\usepackage{multirow}
\usepackage{here} 
\usepackage{relsize}
\usepackage{float}
\usepackage{algorithm}
\usepackage{algcompatible}
\usepackage{caption}
\usepackage{pifont}
\usepackage{amsmath,amsfonts}
\usepackage{array}
\usepackage{textcomp}
\usepackage{stfloats}
\usepackage{verbatim}
\usepackage{cite}
\usepackage{hyperref}

\newcommand{\cmark}{\ding{51}}%
\newcommand{\xmark}{\ding{55}}%

\newcommand{\rebuttal}[1]{{\color{black} #1}}
\newcommand{\rebuttaltwo}[1]{{\color{black} #1}}

\begin{document}

\title{Active Learning of Discrete-Time Dynamics for Uncertainty-Aware Model Predictive Control}

\author{
  Alessandro~Saviolo$^1$,
  Jonathan~Frey$^2$,
  Abhishek~Rathod$^1$,
  Moritz~Diehl$^2$, and
  Giuseppe~Loianno$^1$
\thanks{$^1$The authors are with Tandon School of Engineering, New York University, New York, USA. e-mail: {\tt\footnotesize \{as16054,arathod,loiannog\}@nyu.edu}.}%
\thanks{$^2$The authors are with University of Freiburg, Freiburg, Germany. e-mail: {\tt\footnotesize \{jonathan.frey,moritz.diehl\}@imtek.uni-freiburg.de}.}
\thanks{This work was supported by the NSF CAREER Award 2145277, the DARPA YFA Grant D22AP00156-00, Qualcomm Research, Nokia, and NYU Wireless.}
}

\markboth{IEEE Transactions on Robotics, 2023}%
{Shell \MakeLowercase{\textit{et al.}}: Bare Demo of IEEEtran.cls for IEEE Journals}

\makeatletter
\g@addto@macro\@maketitle{
\setcounter{figure}{0}
\centering
\includegraphics[height=12em]{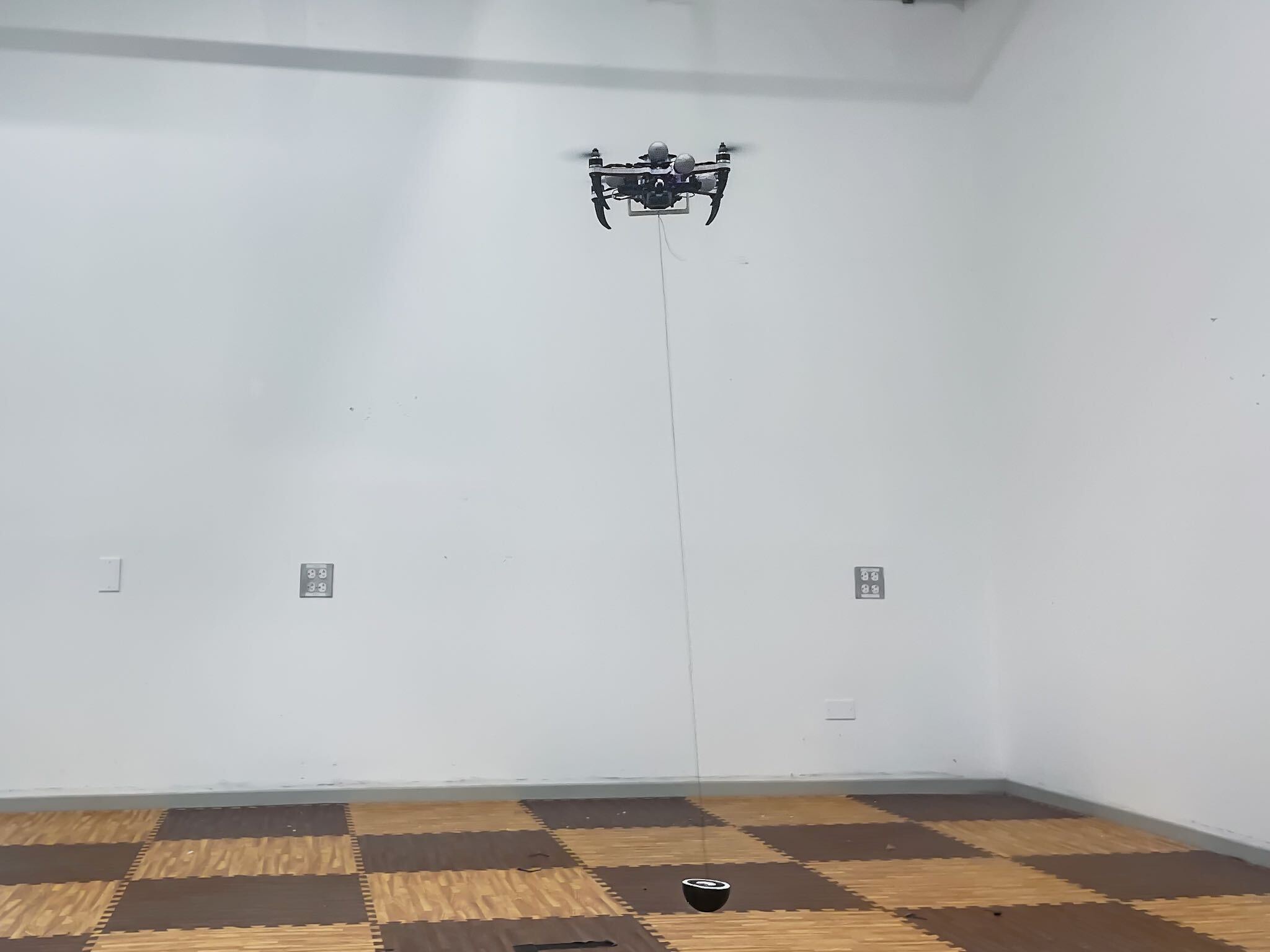}
\includegraphics[height=12em]{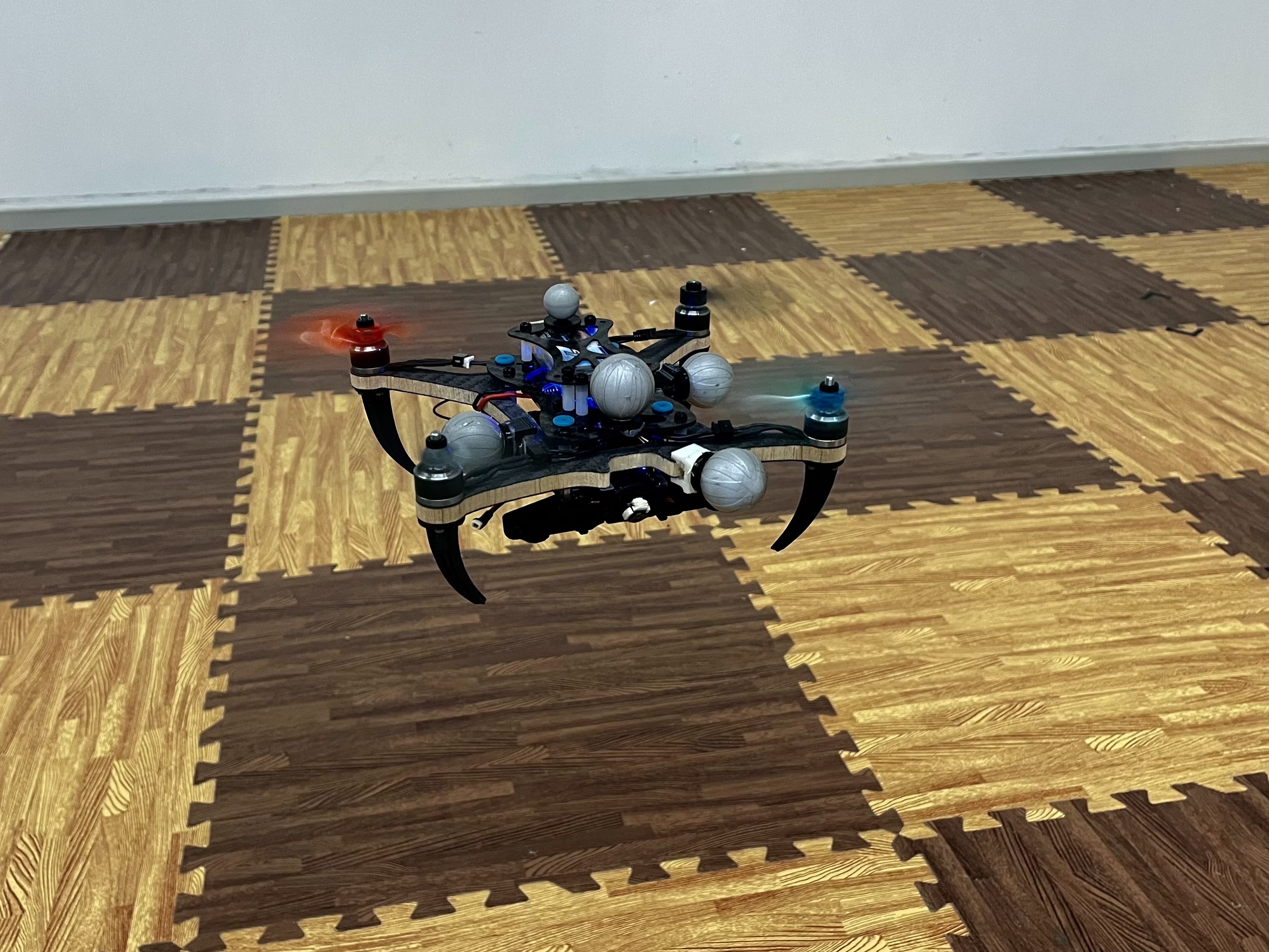}
\includegraphics[height=12em]{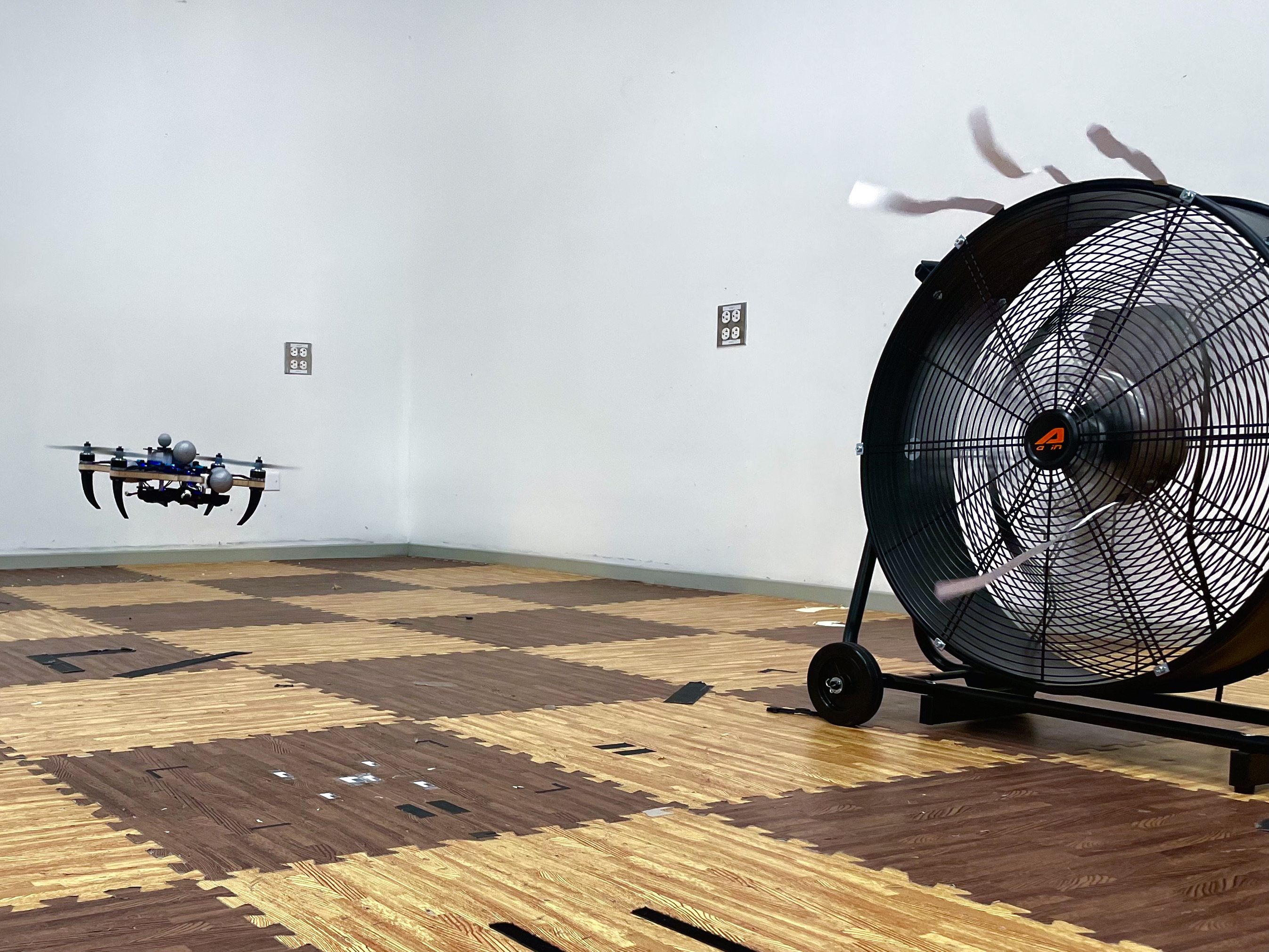}
\captionof{figure}{By combining online learning with uncertainty-aware model predictive control, the learned dynamics actively adapt to multiple challenging operating conditions, enabling unprecedented flight control.\label{fig:initial_figure}}
\vspace{-1.5em}
}
\makeatother
\maketitle

\begin{abstract}
Model-based control requires an accurate model of the system dynamics for precisely and safely controlling the robot in complex and dynamic environments.
Moreover, in presence of variations in the operating conditions, the model should be continuously refined to compensate for dynamics changes.
In this paper, we present a self-supervised learning approach that actively models the dynamics of nonlinear robotic systems. 
We combine \textit{offline} learning from past experience and \textit{online} learning from current robot interaction with the unknown environment.
These two ingredients enable a highly sample-efficient and adaptive learning process, capable of accurately inferring model dynamics in real-time even in operating regimes that greatly differ from the training distribution.
Moreover, we design an uncertainty-aware model predictive controller that is \rebuttal{heuristically} conditioned to the aleatoric (data) uncertainty of the learned dynamics.
This controller \textit{actively} chooses the optimal control actions that (i) optimize the control performance and (ii) improve the efficiency of online learning sample collection.
We demonstrate the effectiveness of our method through a series of challenging real-world experiments using a quadrotor system.
Our approach showcases high resilience and generalization capabilities by consistently adapting to unseen flight conditions, while it significantly outperforms classical and adaptive control baselines.
Video: \url{https://youtu.be/QmEhSTcWob4}
\end{abstract}

\begin{IEEEkeywords}
Model Learning for Control; Aerial Systems: Mechanics and Control; Learning and Adaptive Systems; Optimization and Optimal Control
\end{IEEEkeywords}

%

\IEEEpeerreviewmaketitle

\section{Introduction}
\IEEEPARstart{M}{odel-based} control is an attractive framework for robot control thanks to its adaptivity, scalability, and data efficiency~\cite{harrison2018metarl, chua2018mbrl, williams2017mbrl, moerland2023model}.
\rebuttal{Despite these advantages, the effectiveness of this framework heavily relies on how accurately it can model the robot's interaction with its environment.
For example, developing an accurate model for an aerial vehicle requires capturing complex factors like wind forces, propeller rotation, aerodynamic effects, and vibrations~\cite{saviolo2023learning}. 
Additionally, it is crucial to continually adjust this model to match changes in operating conditions.
For example, the platform may be extended with cable-suspended payloads that would significantly change the dynamics by varying mass and moment of inertia (Figure~\ref{fig:initial_figure}).
Failing to adapt could result in diminished control performance and even catastrophic failures.}

Traditional modeling of the system dynamics is performed using physics-based principles~\cite{yoshida2002roverdynamics, wieber2016leggdynamics, loianno2017quadmodel}.
While these methods can precisely identify rigid-body systems, they fail to capture complex non-linear phenomena, such as friction, deformation, and aerodynamic effects, that cannot be directly measured and \rebuttal{lack} explicit analytic equations.
To address this, recent studies have explored data-driven approaches for modeling robot dynamics using different learning paradigms, from offline~\cite{saviolo2022pitcn, crocetti2023gapt, bauersfeld2021neurobem, duong2021tuninghamilton, punjani2015helicopter, brunton2016discovering, kulathunga2023residual, folkestad2022koopnet, li2022disturbance, an2022fast} to online~\cite{fu2016Oneshotlo, wang2018onlinegp, oconnell2022neuralfly, jiahao2023online} and active learning~\cite{chakrabarty2022activelearning, capone2020activelearning, abraham2019active}.

Offline learning is a setting where the dynamics model is trained on a set of previously collected demonstration data. 
This learning approach ensures both training over the entire data (i.e., capturing a global dynamical model of the system) and learning without posing any physical risk to the robot. 
However, offline learning assumes that the global model will remain accurate over time and thus cannot handle varying operating conditions characteristic of most real-world environments.
Online learning extends offline learning by continuously updating the model during operation with newly collected data, thereby relaxing the need for a perfect system prior. 
However, this learning strategy treats the learned model as a passive recipient of data to be processed, neglecting the system's ability to act and influence the operating environment for effective data gathering.
Active (online) learning corrects this by enabling the robot to take actions that aid its learning process more effectively~\cite{cohn1996activelearning, ren2021survey}. 
The action selection is performed to maximize a performance gain, such as the maximum entropy~\cite{costabal2020activelearning}. 
When the actions are properly selected, the model achieves higher sample efficiency.

In this work, we propose a self-supervised learning framework for actively learning the dynamics of nonlinear robotic systems. 
\rebuttal{Our approach employs a Model Predictive Control (MPC) strategy that harnesses data uncertainty in the learned model to concurrently and efficiently optimize its control performance and learning capability.}
\rebuttal{The specific form of our model, the way we incorporate and combine aleatoric uncertainty, the adaptivity and sample efficiency of our controller in the active learning process as well as the ability to transition this class of approaches from theory to practice and specifically tailored to a resource-constrained, real-time environment such as a small quadrotor are unique aspects of our work compared to existing state-of-the-art solutions.}
Overall, this work presents the following key contributions
\begin{itemize}
    \item Combining \textit{offline} learning from past experience and \textit{online} learning from \rebuttal{current robot-environment interaction while tackling the challenges related to learning stability and robustness inherent in gradient-based online adaptation}. \rebuttaltwo{This combination demonstrates exceptional adaptation and generalization on embedded systems across varied, challenging environments, significantly enhancing the closed-loop control of embedded robotic platforms.}
    \item \rebuttaltwo{Proposing a novel method based on the Unscented Transform (UT) to estimate the aleatoric (data) uncertainty of the learned model, addressing the challenges associated with orientation parametrization from non-Euclidean space of our systems, and capitalizing on its sample and memory efficiency.}
    \item \rebuttal{Designing an MPC, heuristically conditioned by this uncertainty, which actively selects actions to maximize its performance and resilience by efficiently adapting its learned robot model.}
    \item Experimentally validating the proposed methodology to a quadrotor's system, demonstrating successful adaptation of the learned dynamics in multiple challenging real-world experiments. 
    The dynamical model adapts to substantial system configuration modifications (e.g., cable-suspended payload, propeller mixing) and stochastic external effects (e.g., wind disturbance) unseen during training, enabling unprecedented flight control.
\end{itemize}

\section{Related Works}
The proposed learning approach combines offline and active online learning. 
In the following, we discuss these techniques to accurately model and adapt robot dynamics.
\rebuttal{Additionally, we introduce different approaches for modeling uncertainty.}

\textbf{Offline Learning.}
Neural Networks (NNs) have been successfully employed for learning robot dynamics due to their favorable expressive power and flexibility~\cite{saviolo2022pitcn, crocetti2023gapt, bauersfeld2021neurobem, duong2021tuninghamilton, punjani2015helicopter, brunton2016discovering} and the simple self-supervised data collection procedure~\cite{finn2017deepvf, doersch2017ssl}.
Similar to previous works, we seek to learn a NN-based model of the system dynamics purely from robot experience. 
Training the model offline has several advantages compared to pure online learning, such as training without posing any physical risk to the robot and the capability to learn from multiple demonstrations which allows capturing a global model of the system (i.e., the full dynamical evolution of the system).
Most of the NN-based approaches, including those related to quadrotor systems, have focused on learning the continuous-time dynamics in a self-supervised manner~\cite{punjani2015helicopter, bauersfeld2021neurobem, saviolo2022pitcn}.
However, recovering the ground truth translational and rotational accelerations for training the continuous-time NN-based models is extremely complex due to the highly non-linear sensor noise. 
Hence, building accurate datasets for training NN-based models for continuous-time dynamics is extremely difficult and error-prone.
Conversely, in this work, we propose to directly learn the discrete-time dynamics, bypassing the differentiation step required to recover the unobserved accelerations, similarly to~\cite{fu2016Oneshotlo}.
Consequently, the learned dynamical model is trained over less noisy data and can better generalize to unseen data, as shown in the presented experimental results.

Another popular learning scheme for modeling dynamics offline is meta-learning~\cite{finn2017maml, harrison2020alpaca, somrita2020metalearnrover, evans2022iida, nagabandi2019metarl,kaushik2020fast}, which explicitly learns a distribution over different ``tasks'' (e.g., dynamics associated with different operating regimes).
Meta-learning approaches have demonstrated impressive sample efficiency on several tasks, including those considered in this work such as payload transportation~\cite{belkhale2021payloadmetalearn} and wind disturbance adaptation~\cite{oconnell2022neuralfly}.
However, efficiency is largely obtained by integrating a strong inductive bias over the testing task and often requires huge amounts of data for effective training.
Contrarily, the proposed approach is completely task-agnostic, learns offline only a coarse dynamics model from a basic operating regime - alleviating the need for complex hand-designed data sets, and achieves high sample efficiency by actively selecting actions to safely exploit the dynamics model during operation.

\textbf{Online Learning.}
In presence of variations in the operating conditions, the dynamics change, and thus the global model learned offline is no longer accurate.
Online learning is a promising solution for this issue and consists in continuously updating the model during operation as new data is seen~\cite{fu2016Oneshotlo}.
The optimization can be performed over the whole NN model or only the last layers.
Generally, trading-off between these two approaches depends on the similarity of the online and offline training data distributions as well as the available computational budget.
When modeling system dynamics, the two distributions are hardly completely dissimilar but share common representations mostly in the deepest dimensions.
Moreover, continuously refining the whole model would make it less explainable and thus may yield unexpected predictions.
Consequently, in this work, we employ the proposed online learning procedure to only adapt the last layer's weights.

The proposed online optimization procedure is inspired by \cite{bechtle2021forwarderr} and \cite{fu2016Oneshotlo}.
\cite{bechtle2021forwarderr} couples model and controller learning by devising a forward loss similar to our formulation. 
The proposed iterative method allows them to learn an effective control policy on a manipulator and a quadruped in a few iterations. 
However, the approach is validated using basic system configurations in simulation, hence neglecting all the complex effects that make dynamics modeling a challenging task. 
Furthermore, such an iterative method would not be feasible in a real-world setting, particularly for an aerial vehicle such as the quadrotor, due to the extensive time and risk involved in the iterative trial and error process. The fast, non-linear dynamics of quadrotors demand immediate, robust controls, a requirement iterative methods struggle to meet.
We solve these two problems by employing an MPC and combining offline and online learning.
\cite{fu2016Oneshotlo} combines offline and online learning for modeling the dynamics of a manipulator and used MPC for planning under the learned dynamics. 
The general methodology significantly differs from our approach in terms of optimized objectives. 
Moreover, the authors do not consider the aleatoric uncertainty of the learned dynamics to condition the controller. 
Hence, the controller may be overconfident during the adaptation and select poor actions.

Gaussian Processes (GPs) are another attractive class of data-driven models for modeling robot dynamics that can be easily adapted online to new data~\cite{wang2018onlinegp, hewing2020gpmpc}.
However, a major drawback of GP regression is computational complexity. Since these models are non-parametric, their complexity increases with the size of the training set. This implies the need to carefully choose a subset of the fitted data that best represents the true dynamics through data reduction techniques~\cite{sivaram2016fastgp,joaquin2005sparsegp,das2018fastgp}. 
However, since the dynamics are unknown, selecting these points might be challenging.
Moreover, due to their poor scaling properties, GPs need to explicitly model individual dynamical effects (e.g., drag force~\cite{torrente2021datadrivenmpc}).
Generally, the computational complexity of these non-parametric models can be reduced by learning the dynamics component-wise~\cite{sarkka2013spatiotemporal}.
However, this results in the failure to capture the hidden dependencies that bound forces and torques.

Online adaptation resembles adaptive control, which seeks to estimate the model errors online and then reactively account for them at the control level~\cite{joshi2021adaptivecontrol, mao2023propeller, labbadi2019adaptivecontrol, zou2017adaptivecontrol, gahlawat2020l1adap}.
However, this method trades off generalization with performance, which leads to sub-optimal flight performance~\cite{ortega2014l1adap}.
Moreover, in case an MPC is employed as in~\cite{pravitra20201l1adap}, the optimization problem is solved by respecting defective dynamics constraints and thus potentially degrading the controller's predictive performance and accuracy as well as erroneously evaluating the feasibility of additional constraints.
Conversely, in this work, we introduce an online learning procedure for continuously adapting the learned dynamics used by the MPC.
Consequently, the controller fully exploits its predictive power to generate actions, resulting in superior control compared to adaptive control, as shown in the presented experimental results.

\textbf{Active Learning.}
Active learning studies the closed-loop phenomenon of a learner selecting actions that influence what data are added to its training set~\cite{cohn1996activelearning}.
When selecting the actions, the learner may seek to achieve different goals.
Recent works in active learning of robot dynamics have focused on selecting actions that minimize the model (epistemic) uncertainty, hence directing learning gradients toward exploring the entire state-control space.
After a globally accurate model of the system dynamics is obtained, the controller fully exploits the learned model to perform the actual task~\cite{lew2022safeactivelearning}.
While this procedure is well-suited for robots that operate in quasi-static operating regimes, it is not practical for systems with highly-nonlinear dynamics and fast varying operating conditions such as aerial vehicles.
The continuously changing dynamics would never allow the exploration stage to converge.
Moreover, this exploration-exploitation strategy may not reasonably achievable by parametric models (e.g., NNs, which can potentially require an infinite number of points to model the entire state-control space).
In practice, parametric models that are continuously adapted online are not required to fully capture an accurate global model of the system to ensure controllability and stability, but should only be precise locally on a subset of the state-control space.
Therefore, we propose to actively select actions that directly maximize the control performance by conditioning the learner (MPC) to the data (aleatoric) uncertainty.
The conditioned controller continuously exploits the refined dynamical model and chooses ``safe'' actions, resulting in faster convergence of the online learning optimization as demonstrated by the proposed results.

\rebuttal{
\textbf{Uncertainty Estimation}
Learning the dynamics purely from data poses the challenge to make the NN and its training process robust to the aleatoric uncertainty caused by noisy sensor measurements.
Among the key strategies for modeling this uncertainty in NNs are Bayesian inference~\cite{gal2016bayesianinf}, ensemble methods~\cite{valdenegro2019ensemble}, and test time data augmentation~\cite{gawlikowski2021uncertaintysurvey, ebeigbe2021ukf}.
Test time data augmentation methods quantify the predictive uncertainty based on several predictions resulting from different augmentations of the original input sample. 
In contrast to Bayesian and ensemble methods that rely on ad-hoc architectures or multiple NNs to estimate the uncertainty, test time data augmentation approaches require only one NN for both training and inference, do not need changes on the designed NNs, and are not sensitive to the training process initialization parameters.
This motivates us to propose a memory and sample efficient test time data augmentation approach for estimating the dynamics aleatoric uncertainty and heuristically condition the MPC optimization formulation to adapt to the model uncertainties.
}

\begin{figure*}[t]
    \centering
    \includegraphics[width=\textwidth, trim=10 120 100 120, clip]{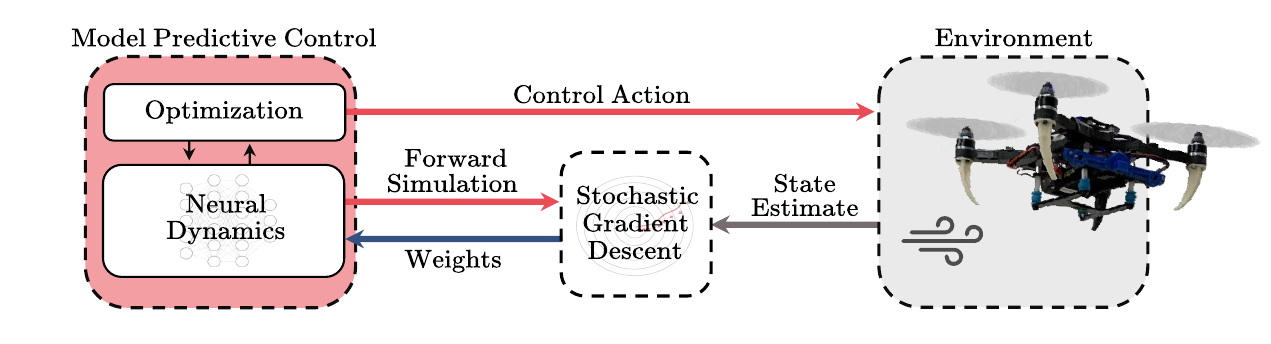}
    \vspace{-1em}
    \caption{Online learning of the system dynamics.
    At each control iteration, the neural dynamics are used by the MPC to calculate the next control action to apply to the system and forward simulated to predict the next state.
    After actuating the predicted action by the controller, the state estimation algorithm provides the actual state reached.
    Finally, the neural dynamics' weights are updated by using the mismatch between the forward simulated state and the actual state reached by the robot.
    \label{fig:online_optimization}}
\end{figure*}

\section{Methodology} \label{sec:methodology}

\subsection{Learning Discrete-Time System Dynamics}
Consider a dynamical system with state vector $\mathbf{x}\in\mathbb{R}^n$ and control input $\mathbf{u}\in\mathbb{R}^m$.
The discrete-time system identification task requires finding a function $f:\mathbb{R}^n \times \mathbb{R}^m \rightarrow \mathbb{R}^n$ such that
\begin{equation*}
    \mathbf{x}_{t+1} = f(\mathbf{x}_{t}, \mathbf{u}_{t})
\end{equation*}
at every time step $t$.
We are interested in solving the system identification problem by approximating $f$ using a feed-forward NN $h(\boldsymbol{\theta})$, where $\boldsymbol{\theta}$ are the NN parameters.
The system identification problem then becomes finding $\boldsymbol{\theta}$ that minimizes the prediction error of $h(\boldsymbol{\theta})$ over a set of demonstration data. 
In this setting, the data consists of state-control trajectories with no additional labels.
Hence, finding $\boldsymbol{\theta}$ is a self-supervised learning task because of the absence of an external supervisory signal (i.e., no human labeling of data~\cite{doersch2017ssl}).

In this work, we provide as input to the NN the translational and rotational velocities, the quaternion representation of the robot orientation, and the control input.
We do not provide position information as input since we assume the dynamics are position-independent. 
We recover the change of position information by using explicit Euler integration~\cite{atkinson1989eulermethod}.
We employ the quaternion representation because it allows a singularity-free mapping from the unit sphere $S^3$ to the group $SO(3)$. 
However, when employing NNs to predict the system dynamics, the predicted orientation has no guarantee to respect the quaternion algebraic constraints (quaternion’s norm can become non-unit, and thus may induce non-orthogonal column vectors in the corresponding rotation matrix).
This may cause undesired behaviors at run-time and would negatively affect the control performance.
Therefore, following~\cite{rucker2018ral}, we treat the predicted orientation $\hat{\mathbf{q}}$ as a non-unit quaternion and map it to a rotation matrix
\rebuttal{
\begin{equation*} \label{app:rucker2018ral}
    \hat{\mathbf{R}} = \frac{\mathbf{Q}}{||\hat{\mathbf{q}}||^2_2}
    ,
\end{equation*}
where
\begin{equation*}
    {\arraycolsep=0.5pt
    \def\arraystretch{1.25}
    \mathbf{Q}
    \resizebox{0.97\columnwidth}{!}{$
    =
    \left[
        \begin{array}{c c c}
        \hat{q}_{w}^{2}+\hat{q}_{x}^{2}-\hat{q}_{y}^{2}-\hat{q}_{z}^{2} & 2\left(\hat{q}_{x} \hat{q}_{y}-\hat{q}_{w} \hat{q}_{z}\right) & 2\left(\hat{q}_{x} \hat{q}_{z}+\hat{q}_{w} \hat{q}_{y}\right) \\
        2\left(\hat{q}_{x} \hat{q}_{y}+\hat{q}_{w} \hat{q}_{z}\right) & \hat{q}_{w}^{2}+\hat{q}_{y}^{2}-\hat{q}_{x}^{2}-\hat{q}_{z}^{2} & 2\left(\hat{q}_{y} \hat{q}_{z}-\hat{q}_{w} \hat{q}_{x}\right) \\
        2\left(\hat{q}_{x} \hat{q}_{z}-\hat{q}_{w} \hat{q}_{y}\right) & 2\left(\hat{q}_{y} \hat{q}_{z}+\hat{q}_{w} \hat{q}_{x}\right) & \hat{q}_{w}^{2}+\hat{q}_{z}^{2}-\hat{q}_{x}^{2}-\hat{q}_{y}^{2}
        \end{array}
    \right]
    $}}
\end{equation*}
}
The matrix $\hat{\mathbf{R}}$ can then be converted to any other orientation representation (e.g., a unit quaternion).

\rebuttal{Using non-unit quaternions can also be advantageous for MPC optimization. Our past experiences with Learning MPC~\cite{li2022learning} suggest that certain integration techniques, such as the Runge-Kutta method, don't fully adhere to the SO(3) structure. Consequently, numerical integration errors may cause the quaternion's norm to drift from unity, resulting in non-orthogonal column vectors within the associated rotation matrix.

While one solution could be to impose a unit constraint (a similar approach might be feasible for the structure NN as well), it introduces further constraints to the optimization problem and thus increases computational complexity. Normalizing the quaternion after each integration might be another alternative, but it's unpractical over an MPC horizon due to the inability to ensure intermediate quaternions within the Runge-Kutta calculations accurately represent SO(3) elements.

In this context, the use of non-unit quaternion representation provides a pragmatic compromise between strict mathematical accuracy and computational efficiency, making it particularly suitable for the MPC problem.}

\subsection{Online Optimization of Learned Dynamics} \label{sec:online_opt}
In presence of variations in the operating conditions, the dynamics change, and the model learned offline is no longer accurate. Therefore, we derive an online optimization strategy that leverages the model learned offline as prior and continuously refines it to the most recent state-control observations.

We control the robot by employing a model-based controller (MPC as detailed in Section~\ref{sec:mpc}) that at each time step calculates the next control action to apply to the system using a dynamics model.
If the model is inaccurate, it will lead to inaccurate predictions internally and degrade the control performance.
To assess the model quality independently of the employed controller, we define the mismatch between the forward simulation of the dynamics model and the next state as \textit{forward error}.
Formally, we define the forward error as
\begin{equation} \label{forward_error}
    \mathcal{L} (\hat{\mathbf{x}}_t, \hat{\mathbf{x}}_{t-1}, \mathbf{u}_{t-1}) = ||  \hat{\mathbf{x}}_t - h(\hat{\mathbf{x}}_{t-1}, \mathbf{u}_{t-1}; \boldsymbol{\theta})||^2_2
    ,
\end{equation}
where $\hat{\mathbf{x}}_t$ is the state estimate at time $t$ and $\mathbf{u}_{t-1}$ is the action applied at the previous time step.
We propose to use the forward error to continuously update the neural dynamics model (Figure~\ref{fig:online_optimization}).
The forward error captures all the inaccuracies of the dynamics and is agnostic to the model-based control policy employed.
Consequently, optimizing the forward error directly translates into adapting the dynamics to the current operating regime.

We calculate at each control iteration the forward error and use stochastic gradient descent~\cite{robbins2007sgd} to update the last layer weights of the NN model to minimize it.
Formally, let $\ell$ be the depth of the NN model, then $\boldsymbol{\theta}$ can be decomposed into $\theta^{1}, \cdots, \theta^{\ell}$.
Consequently, we can formulate the NN such that 
\begin{equation}
h(\mathbf{x}, \mathbf{u}; \boldsymbol{\theta}) = {\theta^{\ell}}^\top \cdot \phi(\mathbf{x}, \mathbf{u}; \theta^{1}, \cdots, \theta^{\ell - 1})
,
\end{equation}
where $\phi(\mathbf{x}, \mathbf{u}; \theta^{1}, \cdots, \theta^{\ell - 1})$ represents a feed-forward NN with depth $\ell - 1$ and $\theta^{\ell}$ is a weight matrix that can be though of as the last linear layer of $h(\mathbf{x}, \mathbf{u}; \boldsymbol{\theta})$.
The weight update is therefore defined as
\begin{equation} \label{eq:online_opt}
    \theta^{\ell}_t = 
    \theta^{\ell}_{t-1} -
    \eta \frac{1}{B} \sum_{i=t-B}^t 
    \nabla \mathcal{L} (\hat{\mathbf{x}}_i, \hat{\mathbf{x}}_{i-1}, \mathbf{u}_{i-1})
    ,
\end{equation}
where $t$ is the current control iteration and $B$ defines the batch size.
The batch size and learning rate are hyper-parameters that define the NN's ability to adapt and should be carefully chosen.
For example, if the environment is significantly affected by external stochastic disturbances or the state estimation is corrupted by noisy spikes, then $B$ should be large to collect more samples while $\eta$ may be small to take short steps.

\rebuttal{
Mini-batch optimization offers several key advantages, including stability, generalization performance, and robustness~\cite{smith2020sgdbenefits}. 
In particular, by providing mini-batches, the updates to the model parameters become less erratic, and the learning process becomes more stable.
Moreover, the stochasticity introduced by mini-batch optimization can help the model to explore different parts of the data distribution and avoid getting stuck in local optima, therefore boosting the generalization performance of the learning process.
Finally, mini-batch optimization can be used to mitigate the effects of noisy data by averaging the gradients computed on different mini-batches, which can help to smooth out the noise and produce more accurate updates to the model parameters.
}

\subsection{Uncertainty Estimation of Neural Dynamics} \label{sec:uncertainty_est}
\rebuttal{Learning system dynamics from collected data introduces the challenge of ensuring that the NN and its training process can handle the aleatoric uncertainty caused by noisy sensor measurements. Inspired by Unscented Kalman Filters, we use the UT~\cite{goderer2003ukfquat} to estimate this uncertainty in the dynamics.}

Denote with $\boldsymbol{\Sigma}$ the corresponding covariance matrix.
Given that the previous mean $\mathbf{\overline{x}}_{t-1}$ and covariance $\boldsymbol{\Sigma}_{t-1}$ matrices are known, the UT is accomplished in three sequential steps:
\begin{enumerate}
    \item Generation of a discrete distribution through sigma points having the same first and second-order moments of the prior data distribution;
    \item Propagation of each point in the discrete approximation through the neural dynamics;
    \item Computation of the current mean $\mathbf{\overline{x}}_{t}$ and covariance $\boldsymbol{\Sigma}_{t}$ from the transformed ensemble.
\end{enumerate}

\rebuttal{When computing the UT, state vectors need to be summed and subtracted, which is not directly possible when using quaternions because they do not belong to Euclidean space.
Therefore, we need to define a procedure for adding and summing quaternions.
Following \cite{wueest2018estimation}, we employ the exponential map ($\exp_{\mathbf{q}}$) with the quaternion parameterization to commute between a minimal orientation parameterization of rotations, defined as the tangent space at the identity of the Lie group $SO(3)$, and a member of this group
\begin{equation}
    \exp_{\mathbf{q}} =
    e^{q_{w}}
    \left[
    \begin{array}{c}
    \cos \left\|\mathbf{q}_{v}\right\| \\
    \frac{\mathbf{q}_{v}}{\left\|\mathbf{q}_{v}\right\|} \sin \left\|\mathbf{q}_{v}\right\|
    \end{array}
    \right],
\end{equation}
where $\arraycolsep=1.25pt \mathbf{q}_v=\begin{bmatrix}q_x&q_y&q_z\end{bmatrix}$ is the vector part of the quaternion.
The logarithm map ($\log_{\mathbf{q}}$) performs the inverse operation
\begin{equation}
    \begin{split}
        \log_{\mathbf{q}} &= 
        \log \left(\|\mathbf{q}\| \frac{\mathbf{q}}{\|\mathbf{q}\|}\right)\\
        &=\log \|\mathbf{q}\|+\log \frac{\mathbf{q}}{\|\mathbf{q}\|}\\
        &=\log \|\mathbf{q}\|+\mathbf{u} \theta\\
        &=\left[\begin{array}{c}
        \log \|\mathbf{q}\| \\
        \mathbf{u} \theta
        \end{array}\right]
        ,
    \end{split}
\end{equation}
where $\mathbf{v} = \mathbf{u} \theta$, with $\theta = ||\mathbf{v}|| \in \mathbb{R}$ and $\mathbf{u}$ unitary.

To simplify the notation, we introduce the two operators $\boxminus, \boxplus$ to perform the operations $-, +$ between the manifold and the tangent space.
We define them as
\begin{equation}
\begin{split}
    \mathbf{q}_1 \boxminus \mathbf{q}_2 &\coloneqq 2 \log _{\mathbf{q}} ( \mathbf{q}_2^* \otimes \mathbf{q}_1 ),\\
    \mathbf{q} \boxplus \boldsymbol{\delta} &\coloneqq \mathbf{q} \otimes \text{exp}_{\mathbf{q}} ( \boldsymbol{\delta} / 2),
\end{split}
\label{app:boxminusplus}
\end{equation}
where $\boldsymbol{\delta}$ is a vector in the tangent space that represents a rotation around the axis of $\boldsymbol{\delta}$ with angle $|\boldsymbol{\delta}|$, $(\cdot)^*$ is the inverse of the quaternion, and $\otimes$ is the quaternion product
\begin{equation}
    \mathbf{p} \otimes \mathbf{q}=\left[\begin{array}{l}
    p_{w} q_{w}-p_{x} q_{x}-p_{y} q_{y}-p_{z} q_{z} \\
    p_{w} q_{x}+p_{x} q_{w}+p_{y} q_{z}-p_{z} q_{y} \\
    p_{w} q_{y}-p_{x} q_{z}+p_{y} q_{w}+p_{z} q_{x} \\
    p_{w} q_{z}+p_{x} q_{y}-p_{y} q_{x}+p_{z} q_{w}
    \end{array}\right].
\end{equation}
Leveraging the map in (\ref{app:boxminusplus}) and given that the dimensionality of $\mathbf{x}$ is $n$, we can now detail the UT methodology.
We use the covariance matrix $\boldsymbol{\Sigma}_{t-1}\in \mathbb{R}^{n\times n}$ and mean $\mathbf{\overline{x}}_{t-1}$ to get a set $\{ \mathcal{X}_{t-1}^{(i)}\}$ of $2n + 1$ \textit{sigma points},
\begin{equation}
\begin{split}
    \mathcal{X}^{(0)}_{t-1} &= \mathbf{\overline{x}}_{t-1},\\
    \mathcal{X}^{(i)}_{t-1} &= \mathbf{\overline{x}}_{t-1} \boxminus \sqrt{n+\lambda} [\sqrt{\boldsymbol{\Sigma}_{t-1}}]_i, \quad i=1, \ldots, n\\
    \mathcal{X}^{(i)}_{t-1} &= \mathbf{\overline{x}}_{t-1} \boxplus \sqrt{n+\lambda} [\sqrt{\boldsymbol{\Sigma}_{t-1}}]_i, \quad i=n+1, \ldots, 2 n
\end{split}
\end{equation}
where $\lambda = \alpha^2 (n + \kappa) - n$ depends on hyper-parameters $\alpha, \kappa$ and determines the sigma points spread, and the square root of the covariance is obtained via Cholesky decomposition.

The state distribution is now represented by a minimal set of carefully chosen sigma points.
These points well capture the mean and covariance of the original distribution and can be forward propagated through the neural dynamics to project them ahead of time
\begin{equation}
    \mathcal{X}_{t}^{(i)} = 
    h( \mathcal{X}_{t-1}^{(i)}, \mathbf{u}_{t-1}; \boldsymbol{\theta}),
    \qquad i=0, \ldots, 2 n
\end{equation}
Finally, the posterior mean and covariance can be accurately calculated by using a weighted sample mean and covariance of the posterior sigma points
\begin{equation}
\begin{array}{l}
    \hat{\boldsymbol{\mu}}_{t} = 
    \mathcal{X}_{t}^{(0)} \boxplus \sum_{i=0}^{2n} 
    W_{i}^{(m)} \biggl(
    \mathcal{X}_{t}^{(i)}
    \boxminus
    \mathcal{X}_{t}^{(0)}
    \biggr),\\
        \hat{\boldsymbol{\Sigma}}_{t} = \sum_{i=0}^{2n} W_{i}^{(c)} 
    \left( \mathcal{X}_{t}^{(i)} \boxminus \hat{\boldsymbol{\mu}}_{t} \right)
    \left( \mathcal{X}_{t}^{(i)} \boxminus \hat{\boldsymbol{\mu}}_{t} \right)^\top,
\end{array}
\end{equation}
where $W_{i}$ is the weight associated with the $i$-th sigma point defined as
\begin{equation}
\begin{array}{l l}
    W_{0}^{(m)} &= \dfrac{\lambda}{n + \lambda}, \\ 
    W_{i}^{(m)} &= \dfrac{1}{2(n+\lambda)},\\ 
    W_{0}^{(c)} &= \dfrac{\lambda}{n + \lambda} + \left( 1 - \alpha^{2} + \beta \right), \\ 
    W_{i}^{(c)} &= \dfrac{1}{2(n+\lambda)} , \\ 
\end{array}
\end{equation}
where $\beta$ is a hyper-parameter that increases the weight of the mean of the posterior sigma points.

The proposed UT methodology for estimating the dynamics uncertainty is memory and sample efficient since it only stores one model and computes $2n+1$ forward evaluations which can be parallelized as in batch training. Moreover, the propagated sigma points capture the posterior mean and covariance accurately up to the 3rd order (Taylor series expansion) for any non-linearity \cite{wan2000ukf}.}

\subsection{Uncertainty-Aware Model Predictive Control} \label{sec:mpc}
In the proposed MPC scheme, we formulate the Optimal Control Problem (OCP) with $N$ multiple shooting steps as
\begin{equation}
\begin{split}
	\min_{\substack{\mathbf{x}_{0},\dots,\mathbf{x}_{N}\\
	\mathbf{u}_{0},\dots,\mathbf{u}_{N-1}}}\quad&
		\frac{1}{2}\sum_{i=0}^{N}
		\Tilde{\mathbf{x}}_i^\top \mathbf{Q}_{\mathbf{x}} \Tilde{\mathbf{x}}_i +
		\frac{1}{2}\sum_{i=0}^{N-1}\Tilde{\mathbf{u}}_i^\top \mathbf{Q}_{\mathbf{u}} \Tilde{\mathbf{u}}_i \\
		\text{s.t.}\quad&~\mathbf{x}_{i+1} = h(\mathbf{x}_{i},\mathbf{u}_{i}; \boldsymbol{\theta}), \quad \textrm{for} \quad i = 0,\dots,N-1\\
		&~\mathbf{x}_0 = \hat{\mathbf{x}}_0 \\
		&~g(\mathbf{x}_{i},\mathbf{u}_{i})\leq 0
	\end{split}
	\label{eq:mpc}
\end{equation}
where $\mathbf{Q}_{\mathbf{x}}$, $\mathbf{Q}_{\mathbf{u}}$ are positive \rebuttal{semi-}definite diagonal weight matrices, while $\Tilde{\mathbf{x}}_i=\mathbf{x}_{\mathrm{des},i} -\mathbf{x}_i$ and $\Tilde{\mathbf{u}}_i=\mathbf{u}_{\mathrm{des},i} -\mathbf{u}_i$ are the errors between the desired state and input and the actual state and input.
\rebuttal{Therefore, the cost function calculates the discrepancy between the predicted and reference states over the time horizon, using multiple reference points.}
The system dynamics are represented by $h(\mathbf{x}_{i},\mathbf{u}_{i}; \boldsymbol{\theta})$ and the initial state is constrained to the current estimate $\hat{\mathbf{x}}_0$.
The problem is further constrained by state and input constraints $g\left(\mathbf{x}_i,\mathbf{u}_i\right)\leq0$ which comprise actuator constraints~\cite{mao2021constraints}. \rebuttal{Additional implementation details are reported in Section~\ref{sec:exp_setup}.}

\rebuttal{
The desired control inputs $\mathbf{u}_{\mathrm{des},i}$ can be obtained from the flat outputs of a differential-flatness planner, as described in~\cite{sun2022diffplan, nguyen2021mpcsurvey}. 
The planner, leveraging the property of differential flatness, designs optimal trajectories in the reduced space of flat outputs. These trajectories are then transformed into the full state and input space due to the unique mapping with respect to the flat outputs, resulting in the desired control inputs.}

\begin{figure*}[t]
    \centering
    \subfigure[Default]{\includegraphics[width=0.25\linewidth, trim=80 0 80 0, clip]{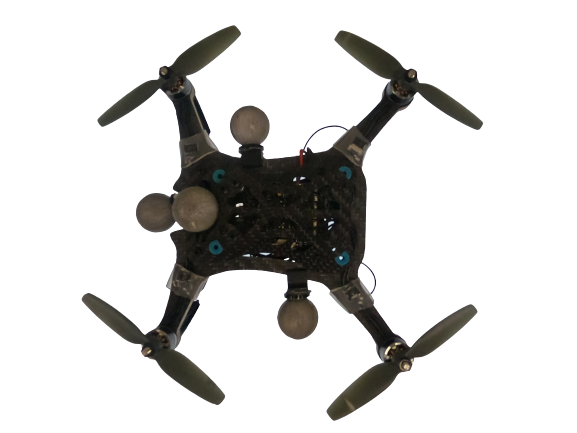}}
    \hspace{2em}
    \subfigure[Payload]{\includegraphics[width=0.25\linewidth, trim=80 0 80 0, clip]{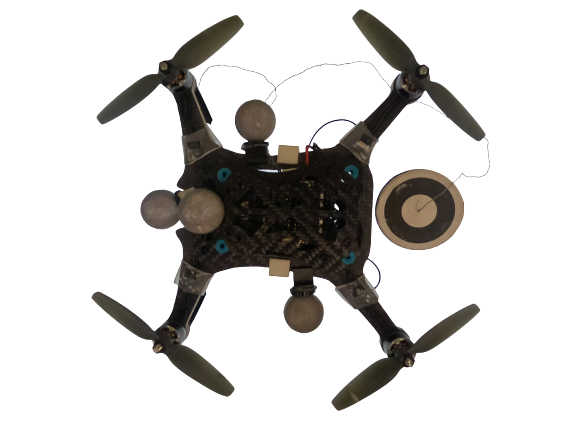}}
    \hspace{2em}
    \subfigure[Mixed Propellers]{\includegraphics[width=0.25\linewidth, trim=80 0 80 0, clip]{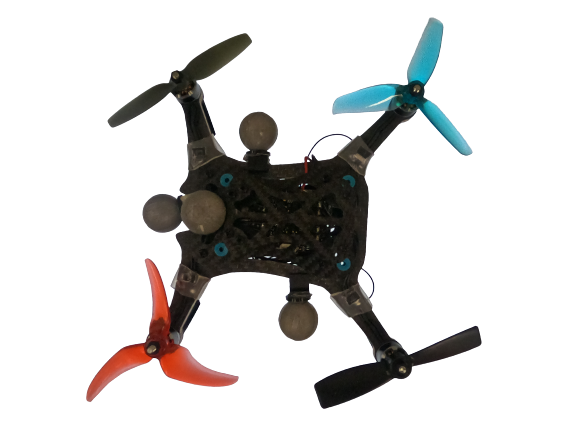}}
    \caption{Quadrotor system configurations used in this work. Payload (b) extends Default (a) with a cable-suspended payload that introduces a static model mismatch due to the mass increase and stochastic effects due to the unknown payload swinging motion. Mixed Propellers (c) critically changes Default (a) by using $4$ different propellers with various blade compositions.}
    \label{fig:system_conf}
\end{figure*}

At each control iteration at time $t$, the mean $\mathbf{\overline{x}}_{t-1}$ and covariance $\boldsymbol{\Sigma}_{t-1}$ matrices of the state distribution are forward propagated in time as described in Section~\ref{sec:online_opt}.
Then, the cost matrix $\mathbf{Q}_{\mathbf{x}}$ in (\ref{eq:mpc}) is adjusted by the neural dynamics uncertainty and updated as
\begin{equation} \label{eq:heuristic}
    \mathbf{Q}_{\mathbf{x}} \leftarrow \mathbf{Q}_{\mathbf{x}}~\mathrm{diag}^{-1} \left(\boldsymbol{\Sigma}_t\right)
    ,
\end{equation}
where $\mathrm{diag}^{-1}$ is the inverse of the matrix diagonal.
\rebuttal{
This adjustment process is essentially a heuristic that reduces the cost associated with deviations from the desired trajectory when the neural network's predictions exhibit higher uncertainty. 
Conversely, the cost increases for deviations when the neural model exhibits higher confidence. 
The outcome is a controller conditioned to favor safer actions that maintain the system closer to the desired state with the least uncertainty.

Once these actions are actuated by the controller, and a state estimate is received from the localization algorithm, the dynamics are subsequently refined to better fit the current operating regime. This refining process creates a continual optimization loop: the controller optimizes actions based on the current dynamic model while the dynamics are refined based on the previous optimal action predicted. This loop facilitates faster convergence and superior sample efficiency in the model training, as evidenced by our experimental results in Section~\ref{sec:adapt_exp}.

Notably, while the controller and the dynamics are both influenced by the aleatoric uncertainty, they do not directly factor in this uncertainty. Instead, it is implicitly accounted for through the update of the state weight matrix in eq.~(\ref{eq:heuristic}).
}

\section{Experimental Setup} \label{sec:exp_setup}
\textbf{System.}
We focus on quadrotor systems for conducting experiments.
We choose quadrotors because they are under-actuated mechanical systems with highly-nonlinear dynamics and fast varying flight conditions, thus they pose unique challenges for the system identification task. 

We learn the dynamical system across various quadrotor configurations. Our default system configuration (Figure~\ref{fig:system_conf}a) is a quadrotor weighing $250~\si{g}$, equipped with a Qualcomm\textsuperscript{\textregistered} Snapdragon\textsuperscript{\texttrademark} board and four brushless motors~\cite{loianno2017quadmodel}.
In addition to the default configuration, we extend the system with a cable-suspended payload weighing $75~\SI{}{\gram}$ and with a cable length of $0.8~\SI{}{\meter}$ (Figure~\ref{fig:system_conf}b). 
This alteration increases the overall system mass by $30 \%$. 
Furthermore, we vary the system configuration by employing $4$ different propellers, each with unique blade compositions (Figure~\ref{fig:system_conf}c).

\textbf{Dynamical Model.}
Following the notation introduced in Table~\ref{tab:notation}, we describe the quadrotor's dynamical system by the state vector $\arraycolsep=1.25pt \mathbf{x}=\begin{bmatrix}\mathbf{p}^\top&\mathbf{v}^\top&\mathbf{q}^\top&\boldsymbol{\omega}^\top\end{bmatrix}\in\mathbb{R}^{13}$ and control input $\arraycolsep=1.25pt \mathbf{u}=\begin{bmatrix}{\Omega}_0&{\Omega}_1&{\Omega}_2&{\Omega}_3\end{bmatrix}\in\mathbb{R}^4$.
Therefore, the system identification task translates into finding a function $f:\mathbb{R}^{13} \times \mathbb{R}^4 \rightarrow \mathbb{R}^{13}$.
We approximate $f$ using a feed-forward NN that consists of 3 dense hidden layers of sizes $64$, $32$, $32$.
Each hidden layer is stacked with an ELU activation function~\cite{clevert2016elu}.
We choose ELU over the more commonly used ReLU because it is continuous, differentiable, and slowly becomes smooth.
In particular, the two activation functions are defined as follows
\begin{equation*}
    \begin{split}
        \text{ReLU}(x) &= \max (0, x),\\
        \text{ELU}(x) &=
        \begin{cases}
            x & \text{if } x > 0 \\
            \alpha (\text{exp} (x) - 1) & \text{if } x \leq 0 \\
        \end{cases}.
    \end{split}
\end{equation*}
where $x$ is the input to the layer, $W, b$ are the weight, bias that define the layer's parameters, and $\alpha$ is a hyper-parameter.
ReLU and ELU are very similar except for negative inputs. 
While ELU becomes smooth slowly, ReLU sharply smoothes.
Hence, ELU is better suited for MPC optimization and makes the OCP in (\ref{eq:mpc}) computationally easier to solve.

The NN input consists of linear velocity $\mathbf{v}$, angular velocity $\boldsymbol{\omega}$, orientation $\mathbf{q}$ and control input $\mathbf{u}$, resulting in a $14 \times 1$ tensor.
The NN output is the discrete-time dynamic state of the quadrotor consisting of velocities and orientation, resulting in a $10 \times 1$ tensor.

\begin{figure*}[t]
    \centering
    \subfigure{\includegraphics[width=0.16\textwidth, trim=40 0 20 110,
    clip]{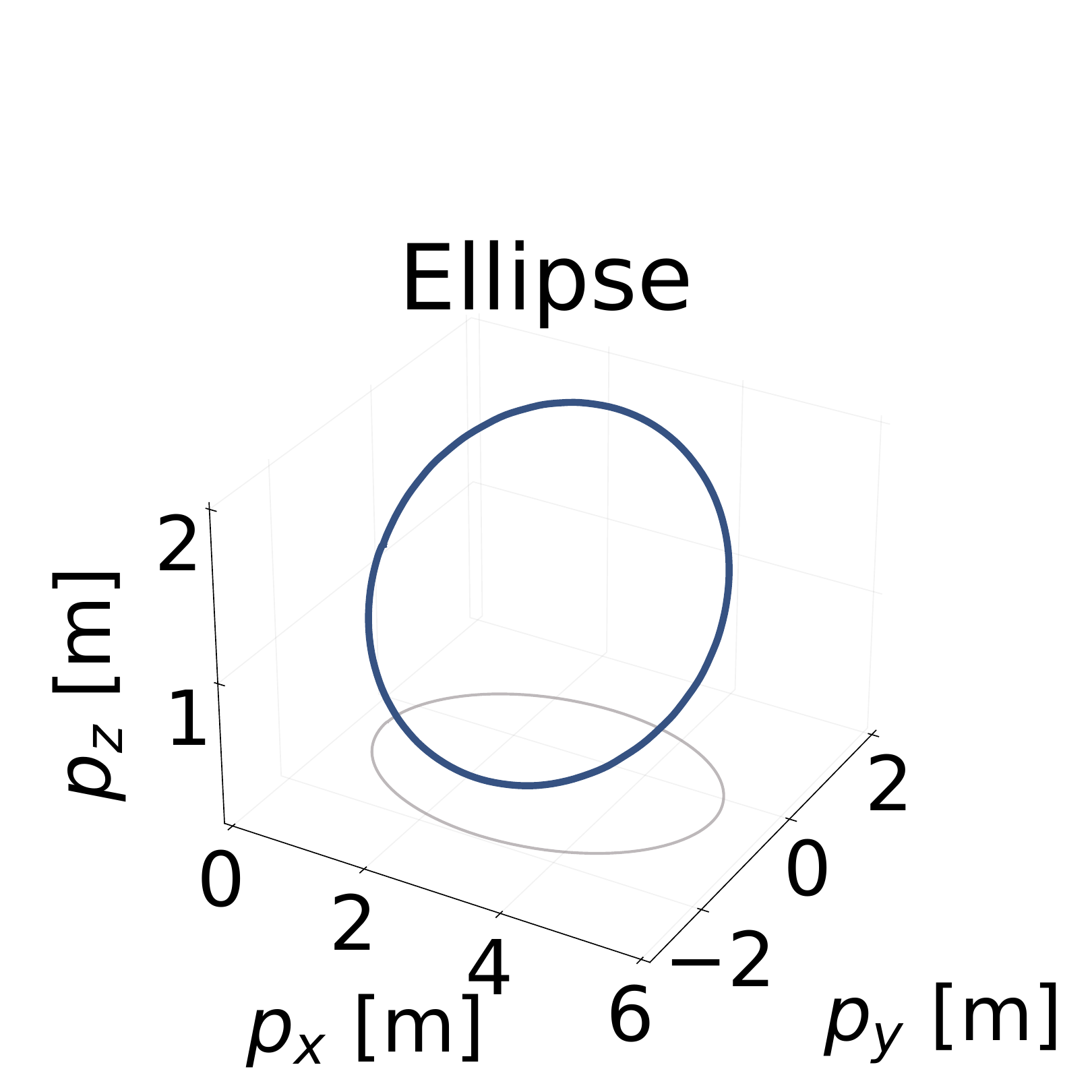}}
    \subfigure{\includegraphics[width=0.16\textwidth, trim=40 0 20 110,
    clip]{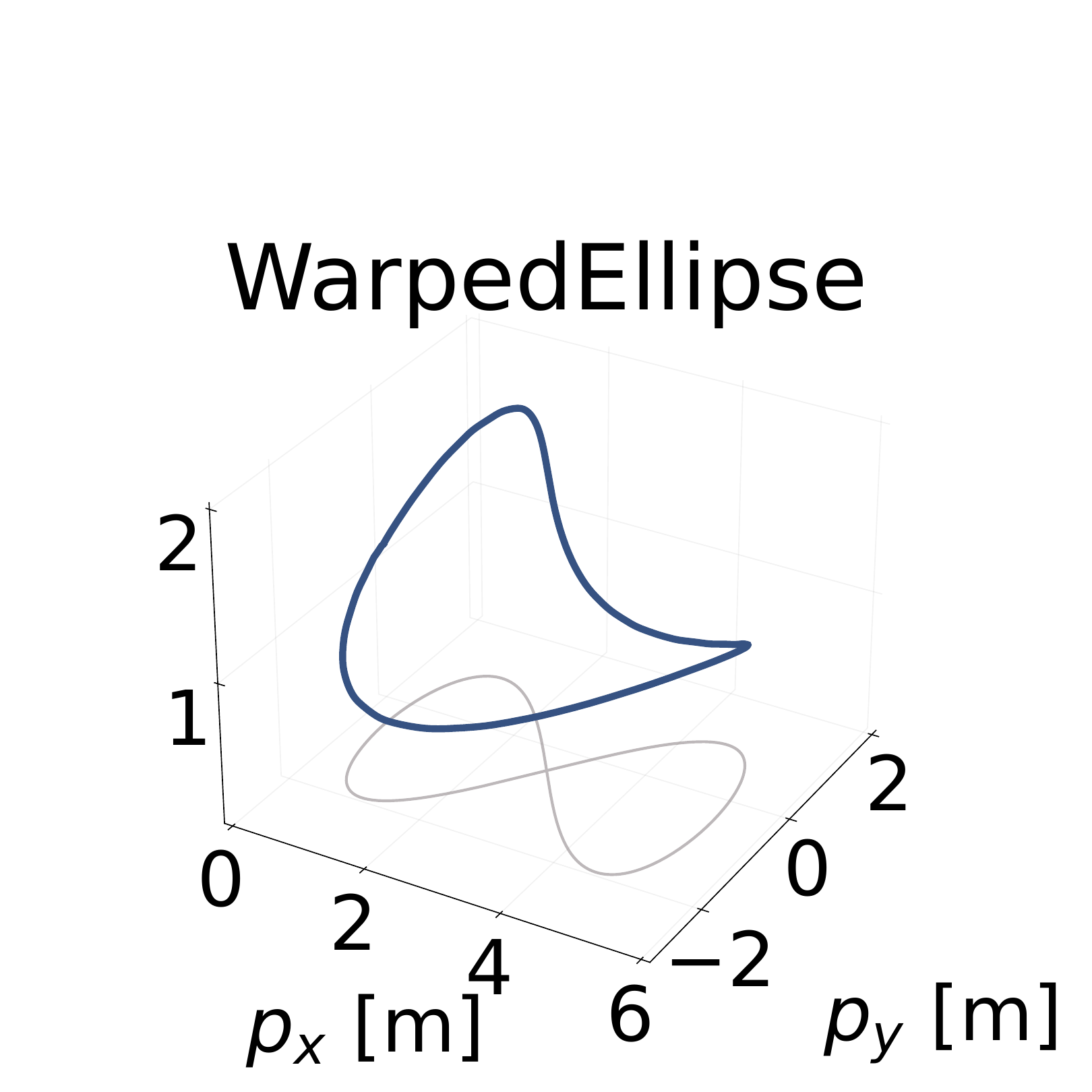}}
    \subfigure{\includegraphics[width=0.16\textwidth, trim=40 0 20 110, clip]{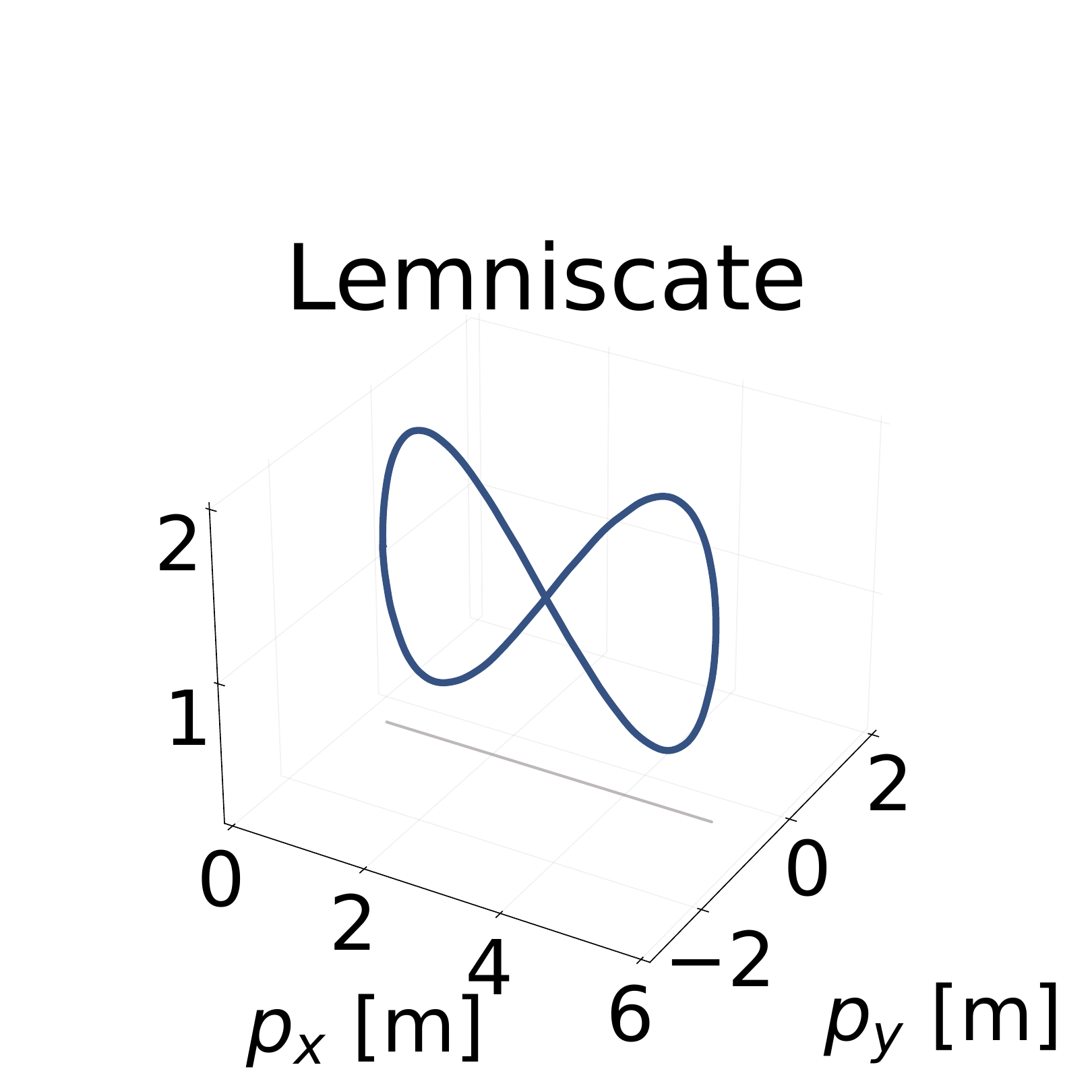}}
    \subfigure{\includegraphics[width=0.16\textwidth, trim=40 0 20 110, clip]{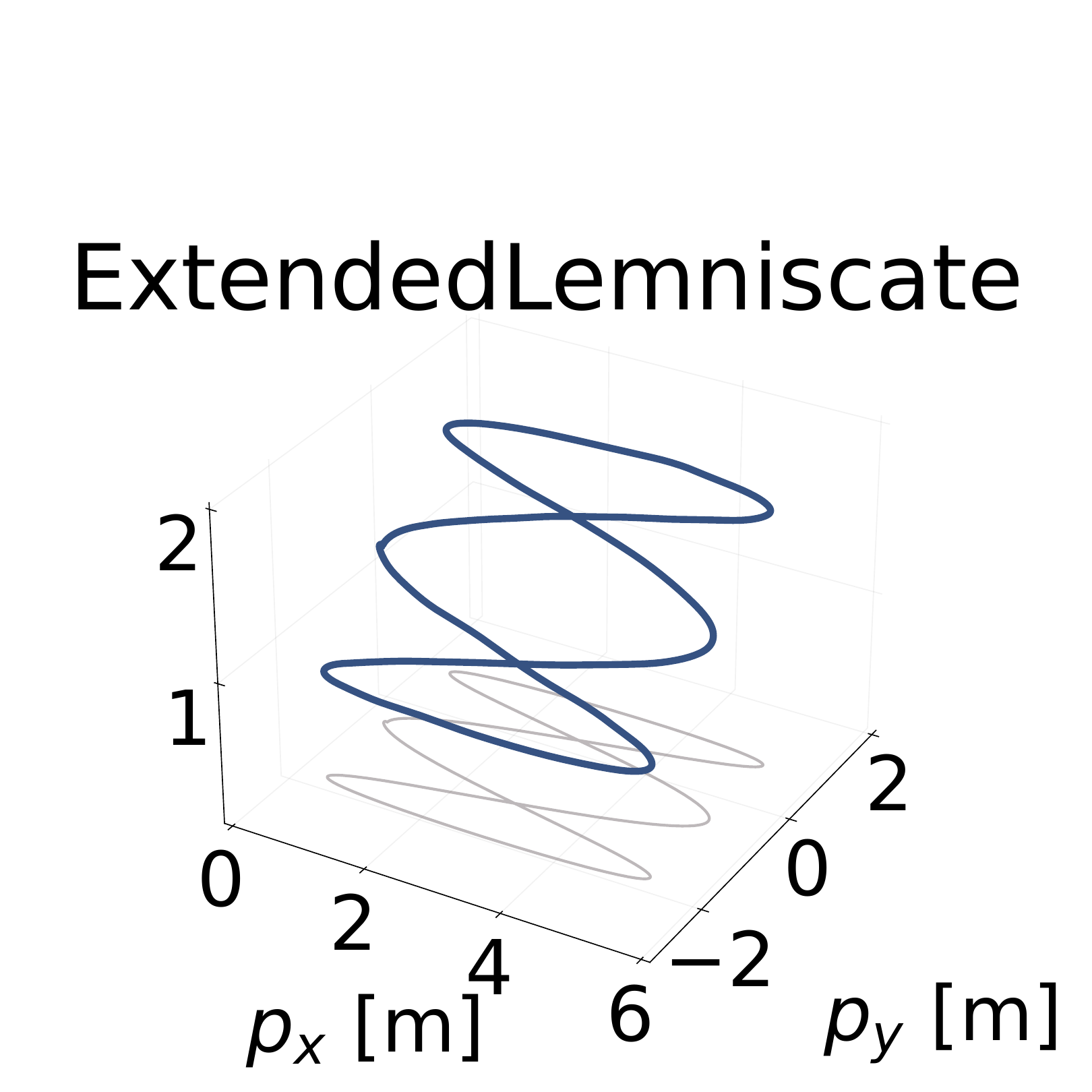}}
    \subfigure{\includegraphics[width=0.16\textwidth, trim=40 0 20 110, clip]{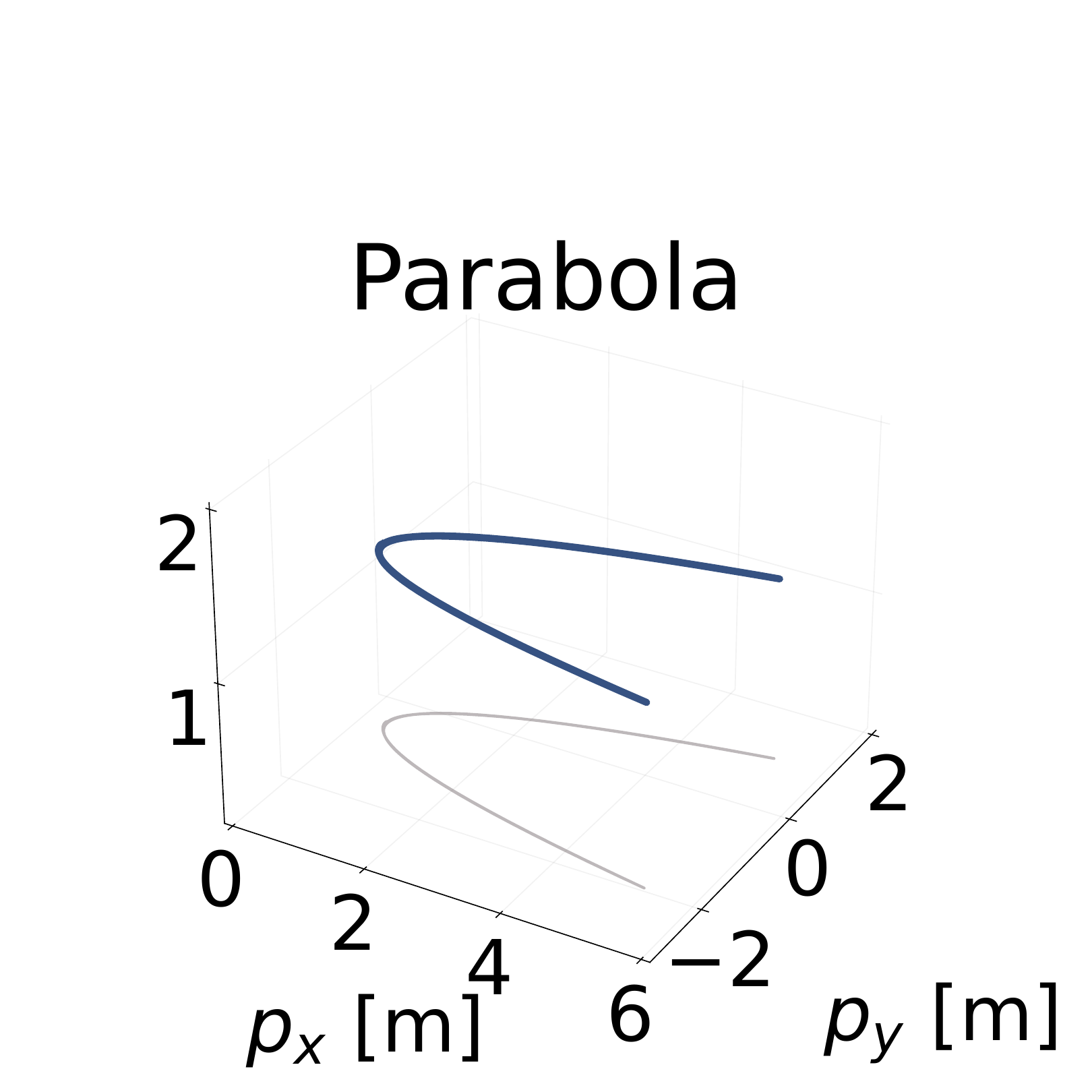}}
    \subfigure{\includegraphics[width=0.16\textwidth, trim=40 0 20 110, clip]{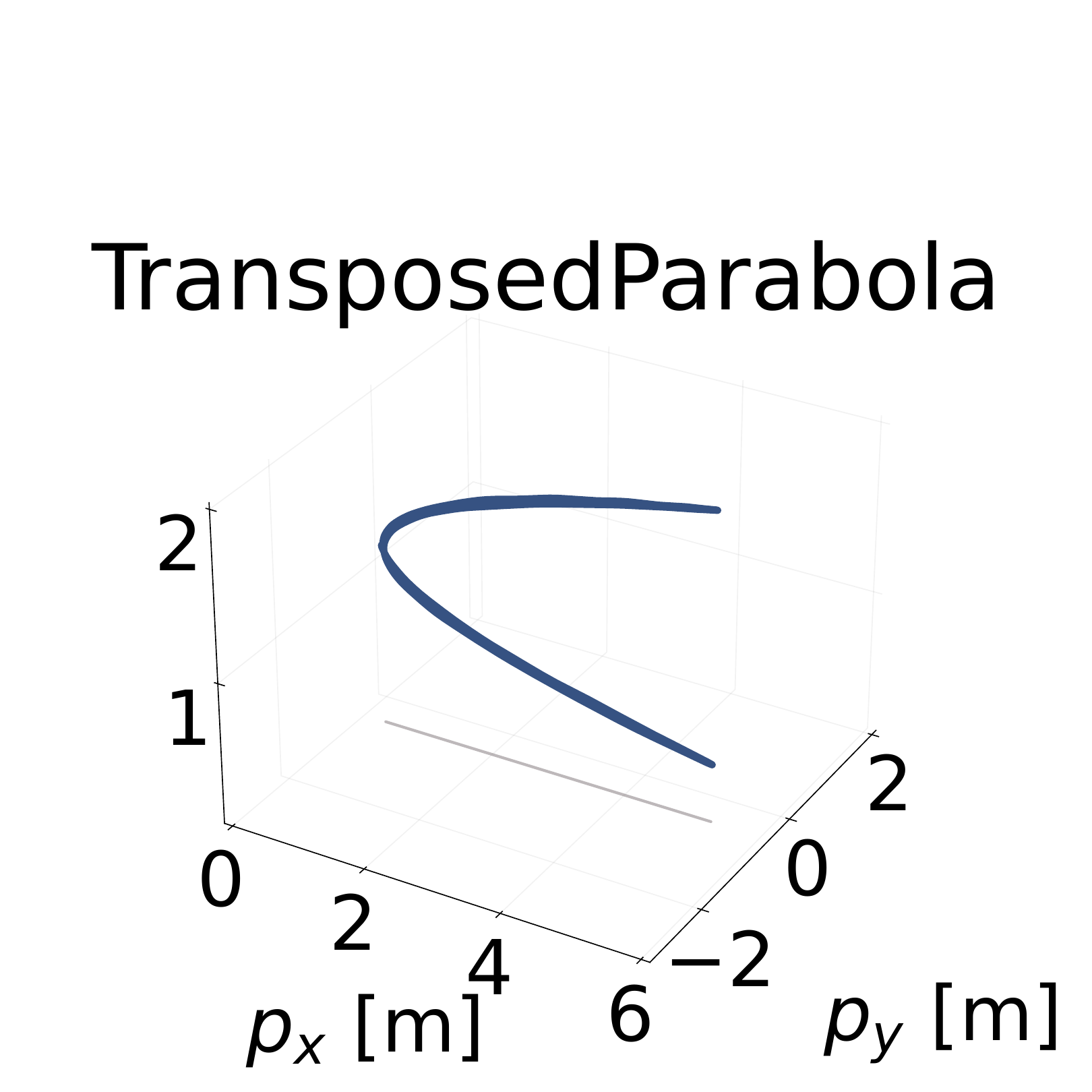}}
    \vspace{-0.5em}
    \caption{Testing trajectories considered in this work. \label{fig:test_trajs}}
    \vspace{-0.5em}
\end{figure*}

For training, we consider the simulated and real-world data sets previously presented and publicly released by \cite{saviolo2022pitcn}.
Each dataset consists of $68$ trajectories with a total of $58^\prime~03^{\prime\prime}$ flight time, where $6$ trajectories are held out for testing (Figure~\ref{fig:test_trajs}).
Based on the results from our initial experiments and analysis, we select one of the two data sets for the subsequent training and testing phases. Further details about the chosen data set for specific experiments can be found in the corresponding subsection. Additionally, a more comprehensive overview of the employed data set is provided in Appendix\ref{app:data_collection} and Appendix\ref{app:pred_perf}.

We run the training process for $20$K epochs using Adam optimizer, batch of $1024$ samples, and learning rate $0.001$. For the online learning phase, we set the learning rate to $0.0002$ and the batch size to $20$.
The training process is performed using PyTorch, while the update of the weights during the online optimization is conducted in a custom implementation of the NN-based dynamics in C++.

\begin{table*}[t]
    \begin{minipage}{.33\linewidth}
        \centering
        \caption{\label{tab:notation}}
        \vspace{-0.75em}
        \caption*{\scshape Notation}
        \begin{tabular}{l l}
            \toprule
            \midrule
            $\mathcal{I}, \mathcal{B}$ & inertial, body frame \\
            $\mathbf{p}\in\mathbb{R}^3$ & position in $\mathcal{I}$ \\
            $\mathbf{v}\in\mathbb{R}^3$ & linear velocity in $\mathcal{I}$ \\
            $\mathbf{q}\in\mathbb{R}^4$ & orientation with respect to $\mathcal{I}$ \\
            $\boldsymbol{\omega}\in\mathbb{R}^3$ & angular velocity in $\mathcal{B}$ \\
            ${\Omega}_i\in\mathbb{R}$ & force produced by the $i$-th propeller \\
            \midrule
            \bottomrule
        \end{tabular}
    \end{minipage}
    \begin{minipage}{.33\linewidth}
        \centering
        \caption{\label{tab:ablation_baselines}}
        \vspace{-0.75em}
        \caption*{\scshape Ablation Baselines}
        \begin{tabular}{c c c}
            \toprule\toprule
            \multirow{2}{*}{Name} & Uncertainty & Online\\
            & Awareness & Learning\\
            \midrule
            Static      & \xmark & \xmark \\
            Static+UA   & \cmark & \xmark \\
            Static+OL   & \xmark & \cmark \\
            Adaptive    & \cmark & \cmark \\
            \bottomrule\bottomrule
        \end{tabular}
    \end{minipage}
    \begin{minipage}{.33\linewidth}
        \centering
        \caption{\label{tab:control_baselines}}
        \vspace{-0.75em}
        \caption*{\scshape Control Baselines}
        \begin{tabular}{c c c}
            \toprule\toprule
            \multirow{2}{*}{Name} & Data-Driven & Adaptive\\
            & Dynamics & Control\\
            \midrule
            Nominal         & \xmark & \xmark \\
            Nominal+L1      & \xmark & \cmark \\
            Static (Ours)   & \cmark & \xmark \\
            Adaptive (Ours) & \cmark & \cmark \\
            \bottomrule\bottomrule
        \end{tabular}
    \end{minipage}
\end{table*}

\textbf{Control.}
We control the quadrotor by running the MPC formulated in Section~\ref{sec:mpc} on a laptop computer \rebuttal{with an Intel Core i7-9750H CPU}.
We send the desired body rates and collective thrust computed by the MPC via Wi-Fi to a PID controller running onboard the quadrotor.

We formulate the MPC OCP in eq.~(\ref{eq:mpc}) with $N=20$ shooting steps, covering the evolution of the system over $1~\si{\second}$.
The optimization is solved using sequential quadratic programming and a real-time iteration scheme~\cite{Diehl2002b} with its implementation in the \texttt{acados} package~\cite{acados}.
The Quadratic Programming (QP) subproblems are obtained using the Gauss-Newton Hessian approximation and regularized with a Levenberg-Marquardt regularization term to improve the controller robustness.
The QPs are solved using the high-performance interior-point method in \texttt{HPIPM}~\cite{Frison2020a} with full condensing and the basic linear algebra library for embedded optimization \texttt{BLASFEO}~\cite{Frison2018}.

The OCP solver has been prototyped using the \texttt{acados} Python interface and the problem functions are generated using \texttt{CasADi}~\cite{casadi}.
However, since CasADi builds a static computational graph and postpones the processing of the data, it is not compatible with PyTorch which directly performs the computations using the data. Hence, the learned dynamics using PyTorch can not be directly embedded in the formulated MPC. We solve this issue by implementing our customized version of NN-based dynamics directly in CasADi symbolics.

\rebuttal{
The OCP in eq.~(\ref{eq:mpc}) is weighted by the definite diagonal weight matrices
\begin{equation*}
    \begin{split}
        \mathbf{Q}_{\mathbf{p}} &= \mathrm{diag}(200, 200, 300) \hspace{0.3em},\\
        \mathbf{Q}_{\mathbf{v}} &= \mathrm{diag}(10, 10, 10) \hspace{0.3em},\\
        \mathbf{Q}_{\mathbf{q}} &= \mathrm{diag}(150, 150, 200, 1) \hspace{0.3em},\\
        \mathbf{Q}_{\boldsymbol{\omega}} &= \mathrm{diag}(10, 10, 10) \hspace{0.3em},\\
        \mathbf{Q}_{\mathbf{u}} &= \mathrm{diag}(5, 5, 5, 5) \hspace{0.3em},
    \end{split}
\end{equation*}
where $\mathrm{diag}$ denotes a diagonal matrix and is formulated to respect the state and input constraints
\begin{equation*}
    \begin{split}
        -10 \leq ~ &\mathbf{v}_i \leq 10 \hspace{2.7em} \forall i \in [0, 2] 
        \hspace{0.5em} [\SI{}{\meter\per\second}],\\
        -3.14 \leq ~ & \boldsymbol{\omega}_i \leq 3.14 \hspace{1.8em} 
        \forall i \in [0, 2] \hspace{0.5em} [\SI{}{\radian\per\second}],\\
        0.01 \leq ~ & \boldsymbol{\Omega}_i \leq 1.124 \hspace{1.25em}
        \forall i \in [0, 3] \hspace{0.5em}  [\SI{}{\newton}].
    \end{split}
\end{equation*}

The UT methodology depends on the choice of the hyper-parameters $\alpha$, $\kappa$, and $\beta$.
which are selected based on the characteristics of the state and measurement noise and the nonlinearity of the system.
In this work, we set $\alpha=0.001$, $\kappa=1.0$, and $\beta=2.0$ assuming a Gaussian prior distribution.
}

\rebuttaltwo{\textbf{Localization.}
We conduct the experiments in an environment equipped with Vicon motion capture system. This choice ensures that the experimental results precisely evaluate the dynamics and controller's accuracy. Introducing onboard sensors presents challenges like sensor noise and state estimation drifts, which would significantly affect tracking performance and, in turn, the demonstration of our contributions.}

\textbf{Metrics.}
We compare our approach against multiple baselines for dynamics learning and quadrotor control based on the Root Mean Squared Error (RMSE) and the Cumulative RMSE (CRMSE).
The error metrics between the true state $\mathbf{x}$ and the predicted state $\hat{\mathbf{x}}$ are formulated as
\begin{equation*}
    \begin{split}
         \text{RMSE}(\mathbf{x}, \hat{\mathbf{x}}) &= 
         \sqrt{\frac{1}{N} \sum _{i=0} ^{N-1} (\mathbf{x}_i - \hat{\mathbf{x}}_i)^\top (\mathbf{x}_i - \hat{\mathbf{x}}_i)},\\
         \text{CRMSE}(\mathbf{x}_t, \hat{\mathbf{x}_t}) &=
         \sum _{i=0} ^{t} \text{RMSE}(\mathbf{x}_i, \hat{\mathbf{x}_i}),
    \end{split}
\end{equation*}
where $N$ is the number of components of the state and $t$ is the number of control iterations from the start of the experiment.

The CRMSE allows us to easily analyze the trend of the error function over a certain period, hence it is a well-suited metric for understanding the role of methods that make a continuous optimization over time.

\begin{table*}[t]
\centering
\rebuttal{
\caption{\label{tab:prediction_error}}
\vspace{-0.75em}
\caption*{\scshape Predictive Performance on Unseen Testing Trajectories}
\begin{tabular}{c c c c c c c c c}
\toprule\toprule
\multirow{2}{*}{Trajectory} & 
$\dot{\mathbf{v}}_{\max}$ &
$\dot{\boldsymbol{\omega}}_{\max}$ &&
\multicolumn{2}{c}{$\mathbf{v}$~RMSE~$[\SI{}{\meter\per\second}]$} && 
\multicolumn{2}{c}{$\boldsymbol{\omega}$~RMSE~$[\SI{}{\radian\per\second}]$} \\
\cline{5-6}\cline{8-9}
& $[\SI{}{\meter\per\second\squared}]$ & 
$[\SI{}{\radian\per\second\squared}]$ && 
Continuous-Time \cite{saviolo2022pitcn} & 
Discrete-Time (Ours) &&
Continuous-Time \cite{saviolo2022pitcn} & 
Discrete-Time (Ours) \\
\midrule
Ellipse
& $ 5.89           $ & $ 7.34           $ && 
  $ 0.097 \pm 0.01 $ & $ \textbf{0.078} \pm \textbf{0.01} $ && 
  $ 0.194 \pm 0.03 $ & $ \textbf{0.089} \pm \textbf{0.02} $\\
WarpedEllipse
& $ 9.99           $ & $ 12.76          $ && 
  $ 0.122 \pm 0.01 $ & $ \textbf{0.109} \pm \textbf{0.01} $ && 
  $ 0.215 \pm 0.07 $ & $ \textbf{0.137} \pm \textbf{0.03} $\\
Lemniscate
& $ 13.14          $ & $ 31.98          $ && 
  $ 0.129 \pm 0.04 $ & $ \textbf{0.111} \pm \textbf{0.05} $ && 
  $ 0.287 \pm 0.14 $ & $ \textbf{0.140} \pm \textbf{0.09} $\\
ExtendedLemniscate
& $ 7.76           $ & $ 27.95          $ && 
  $ 0.099 \pm 0.04 $ & $ \textbf{0.064} \pm \textbf{0.03} $ && 
  $ 0.170 \pm 0.13 $ & $ \textbf{0.102} \pm \textbf{0.03} $\\
Parabola
& $ 5.91           $ & $ 6.57           $ && 
  $ \textbf{0.032} \pm \textbf{0.01} $ & $ 0.065 \pm 0.01 $ && 
  $ 0.102 \pm 0.04 $ & $ \textbf{0.099} \pm \textbf{0.02} $\\
TransposedParabola
& $ 17.86          $ & $ 54.90          $ && 
  $ 0.141 \pm 0.03 $ & $ \textbf{0.107} \pm \textbf{0.04} $ && 
  $ 0.398 \pm 0.22 $ & $ \textbf{0.147} \pm \textbf{0.04} $\\
\bottomrule\bottomrule
\end{tabular}
\vspace{-1.5em}
}
\end{table*}

\section{Results}
We design our evaluation procedure to address the following questions.
(i) How does the predictive performance of our proposed discrete-time model compare with a continuous-time baseline~\cite{saviolo2022pitcn} when tested on real-world data?
(ii) What advantages does the active learning of dynamics present when controlling the quadrotor?
(iii) Can the proposed approach adapt to challenging flight regimes and how does it compare against classical~\cite{loianno2017quadmodel} and adaptive control baselines~\cite{wu2022l1adap}?
\rebuttal{(iv) How can we statistically demonstrate the convergence of the proposed framework?}
We encourage the reader to consult the supplementary material at the end of this paper and video for additional results.

\subsection{Discrete vs. Continuous-Time Dynamics} \label{sec:pred_perf}
We compare the predictive performance of the proposed dynamics model and a continuous-time baseline. 
Both models are trained and tested on the real-world dataset (Figure~\ref{fig:test_trajs}) 
The continuous-time baseline corresponds to the PI-MLP model proposed in~\cite{saviolo2022pitcn}, which was demonstrated to generate more precise predictions compared to multiple baselines for the quadrotor's system identification task, from traditional approaches to learning-based methods.
We feed the continuous-time baseline with the same input information as the proposed discrete-time approach. 
The baseline outputs the continuous-time dynamics of the quadrotor's system, i.e. linear acceleration $\dot{\mathbf{v}}$ and angular acceleration $\dot{\boldsymbol{\omega}}$.
Then, we discretize the continuous-time solutions by employing the Runge-Kutta 4th-order numerical integration.

Table~\ref{tab:prediction_error} summarizes the results of this experiment. 
Both models demonstrate excellent performance over all the proposed maneuvers, capturing all the complex non-linear effects, and performing accurate predictions. 
However, despite being trained on the same dataset and sharing a quasi-identical architecture, the proposed model consistently and significantly outperforms the continuous-time baseline, improving the prediction error by up to $50\%$.
The key reason for this dominant performance is the dynamics' level of abstraction that the model learns. 
Specifically, learning continuous-time dynamics requires explicitly extracting linear and angular accelerations from flight data. However, this usually simply translates into computing the first-order derivative of velocities that are corrupted by noisy measurements. Hence, the computed accelerations are affected by noisy spikes that limit the learning capabilities of the neural model. Low-pass filters can be applied to minimize the noise in the collected data. However, there is no guarantee that the filters' outputs are the true dynamics and that the discarded information is mere noise. Moreover, using filters introduces time delays that may render online learning not effective. Conversely, learning discrete{-time} dynamics does not require any velocity differentiation and therefore the NN is trained to regress less noisy labels.

\begin{table}[t]
    \centering
    \rebuttal{
    \caption{\label{app:uncertainty_benefits}}
    \vspace{-0.75em}
    \caption*{\scshape Closed-Loop Tracking Performance in Simulation}
    \begin{tabular}{c c c c}
        \toprule\toprule
        \multirow{2}{*}{Trajectory} & Static && Static+UA\\
        \cline{2-2}\cline{4-4}
        & $\mathbf{p}$~RMSE~$[\SI{}{\meter}]$ && $\mathbf{p}$~RMSE~$[\SI{}{\meter}]$\\
        \midrule
        Ellipse & $0.34 \pm 0.10$ && $\textbf{0.29} \pm \textbf{0.08}$ \\
        WarpedEllipse & $0.23 \pm 0.10$ && $\textbf{0.21} \pm \textbf{0.12}$ \\
        Lemniscate & $0.14 \pm 0.09$ && $\textbf{0.12} \pm \textbf{0.06}$ \\
        ExtendedLemniscate & $0.09 \pm 0.06$ && $\textbf{0.09} \pm \textbf{0.05}$ \\
        Parabola & $0.28 \pm 0.12$ && $\textbf{0.20} \pm \textbf{0.14}$ \\
        TransposedParabola & $0.17 \pm 0.06$ && $\textbf{0.14} \pm \textbf{0.06}$ \\
        \bottomrule\bottomrule
    \end{tabular}
    }
\end{table}

\begin{figure*}[t]
    \centering
    \includegraphics[width=0.7\linewidth, trim=0 280 0 20, clip]{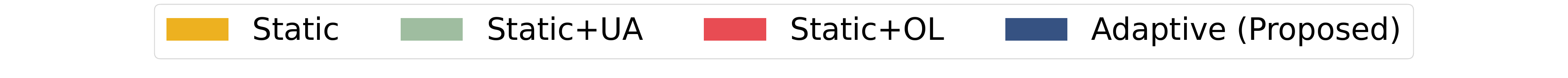}
    \includegraphics[width=0.95\linewidth, trim=0 0 0 20, clip]{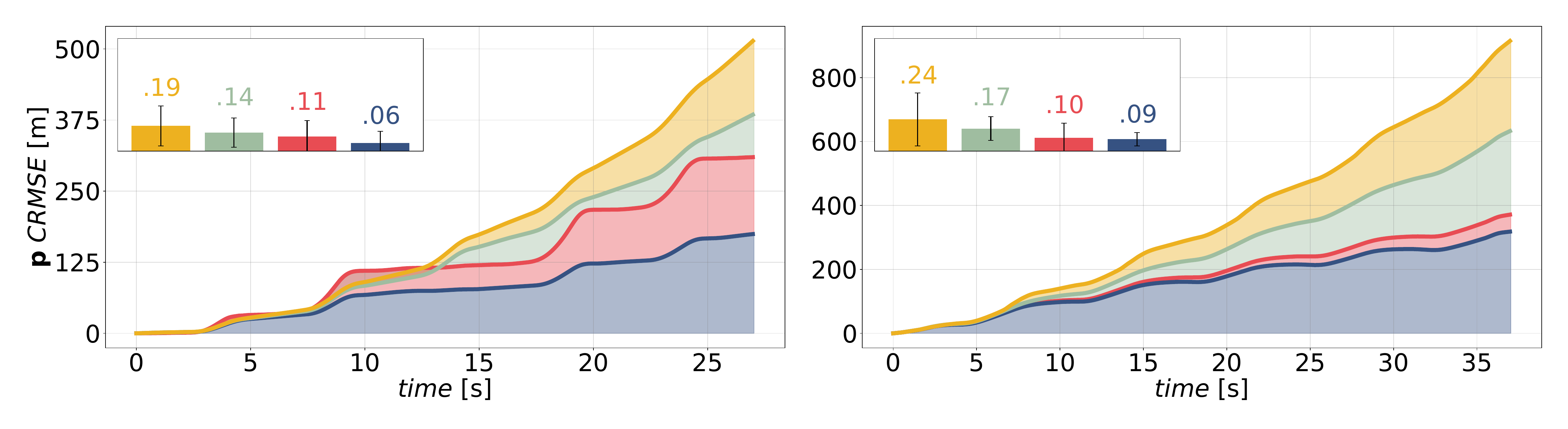}
    \hspace{1em}
    \caption{Benefits of combining online learning with uncertainty estimation when continuously tracking WarpedEllipse (left) and Lemniscate (right) trajectories in the real world.
    The RMSE is accumulated over multiple loops to validate the increased stability introduced by the uncertainty estimation and online optimization. 
    (Inset) Average RMSE over the entire flight performance.
    \label{fig:uncertainty_ablation}}
\end{figure*}

\subsection{Leveraging Uncertainty for Control and Dynamics Learning} \label{sec:uncertainty_benefits}
We demonstrate the benefits of conditioning the MPC objective function on aleatoric uncertainty for controlling the quadrotor and boosting the online learning convergence.
We utilize the MPC, as outlined in Section~\ref{sec:mpc}, to control the quadrotor in order to continuously track multiple testing trajectories in both simulated and real-world settings.
We compare the tracking performance when selectively omitting the proposed Uncertainty-Awareness heuristic (\textit{UA}) and Online Learning strategy (\textit{OL}), resulting in four different control methods as reported in Table~\ref{tab:ablation_baselines}.
For these experiments, the dynamics model is trained on the real-world dataset, even for the flights in simulation.

Table~\ref{app:uncertainty_benefits} provides the mean and standard deviation of the positional RMSE while tracking multiple testing trajectories in simulation.
The results reveal that incorporating the estimated uncertainty into the MPC optimization function improves the stability of the flight controller. 
As a result, the uncertainty-aware MPC \rebuttal{(\textit{Static+UA})} achieves better accuracy and precision in tracking performance compared to the conventional MPC \rebuttal{(\textit{Static})}, reducing the positional RMSE by up to $30 \%$. 
Moreover, error plots show better tracking performance over time for the uncertainty-informed MPC compared to the classical MPC.

In order to provide a fair comparison between uncertainty-aware and uncertainty-unaware MPCs, we use a simulation environment. 
However, this environment lacks aerodynamic forces, motor dynamics, communication delays, and other non-linear effects present in real-world scenarios.
This results in the aleatoric uncertainty of the dynamical model remaining moderate, hence tracking performance is comparable between the uncertainty-aware and uncertainty-unaware MPC during the early stages of flight.
As the flight continues, cumulative tracking errors compound, degrading overall performance. In these later stages, the uncertainty-awareness procedure proves critical for maintaining performance.

The results of similar experiments conducted in the real world are shown in Figure~\ref{fig:uncertainty_ablation}.
These findings underscore the importance of our proposed online learning procedure for continuously improving the dynamics model over time, facilitating high-performance control on aggressive trajectories not previously seen.
When the online learning procedure is enabled \rebuttal{(\textit{Static+OL}, \textit{Adaptive})}, the quadrotor's tracking performance is improved by up to $60\%$.
Furthermore, the results indicate that conditioning the MPC objective function using the estimated uncertainty enables faster model learning convergence and sample efficiency, hence better tracking performance over time compared to an uncertainty-unaware controller.
When the online learning procedure is enabled and the MPC is conditioned \rebuttal{(\textit{Adaptive})} - the proposed active learning method, the tracking performance of the quadrotor is dramatically improved by up to $70\%$.
The primary reason for this is that when the MPC is uncertainty-informed, it opts for safer actions (i.e., prioritizes tracking certain state components), favoring the exploitation of the dynamics model over exploration.
This, in turn, significantly boosts the convergence and sample efficiency of the online model training.

\subsection{Adaptation to Challenging Operating Conditions} \label{sec:adapt_exp}
We evaluate our proposed approach (\textit{Adaptive}) against multiple baselines for quadrotor control in several challenging real-world flight operating conditions, which involve extensive modifications to the system configuration or external disturbances.
Specifically, we consider an MPC that leverages a nominal model of the quadrotor as in~\cite{loianno2017quadmodel} (\textit{Nominal}) and the nominal MPC augmented by an L1 adaptive controller as in~\cite{wu2022l1adap} (\textit{Nominal+L1}).

\textit{Nominal} baseline models the system's dynamics through non-linear differential equations and applies the MPC defined in Section~\ref{sec:mpc} to control the quadrotor. 
However, it falls short in representing complex non-linear phenomena such as friction, deformation, aerodynamic impacts, and vibrations. 
To overcome this limitation, we consider a second baseline: an adaptive controller. This controller augments the MPC's predictions in real time by modeling the complex non-linear phenomena that the nominal model fails to capture. 
We specifically focus on the L1 adaptive controller recently proposed in~\cite{wu2022l1adap} as it has been effectively utilized for quadrotor control and its code is publicly available. 

We keep the parameters the same as the original implementation of the L1 adaptive controller.
For more details on the comparison standards, please may refer to the corresponding papers.
Table~\ref{tab:control_baselines} provides an overview of the different control baselines used in this set of experiments.

We use the controllers mentioned above to control our quadrotor in the following flight operating conditions
\begin{itemize}
    \item \textit{Payload transportation}: 
    learning the dynamics purely from data poses the challenge to make the NN generalize to domain shifts between training and testing distributions. 
    We study the capability of the proposed approach to adapt to significant changes in the vehicle's mass by attaching to the quadrotor's body a cable-suspended payload with mass $75~\SI{}{\gram}$ and cable length $0.8~\SI{}{\meter}$ (Figure~\ref{fig:initial_figure}a).
    The quadrotor is required to take off and track a circular trajectory of radius $1~\SI{}{\meter}$. 
    Note that the payload not only increases the mass of the quadrotor by $30 \%$, but it also introduces stochastic disturbances due to the payload swing motions;
    \item \textit{Mixing propellers}: 
    propellers provide lift for the quadrotor by spinning and creating airflow. Mixing different propellers may cause the system to lose stability and fail to track even simple near-hovering trajectories~\cite{bristeau2009propellers}. Therefore, we validate the adaptability of the proposed approach to significant domain shifts by altering the quadrotor's propellers. Specifically, we significantly change the system configuration by using $4$ different propellers with various blade compositions (Figure~\ref{fig:initial_figure}b). The quadrotor is required to take off and hover at $0.5~\SI{}{\meter}$ height;
    \item \textit{Wind disturbance}: 
    accurate tracking under challenging wind conditions is difficult due to the stochastic nature of wind and the domain shift caused by the new environment dynamics never seen during training. We analyze the adaptability of the proposed approach when an industrial fan blows towards the quadrotor with air speeds up to $3~\SI{}{\meter\per\second}$ (see Figure~\ref{fig:initial_figure}c). The quadrotor is required to take off and hover at $0.5~\SI{}{\meter}$ height.
\end{itemize}

\begin{figure*}
    \centering
    \includegraphics[width=0.9\linewidth, trim=0 0 0 70, clip]{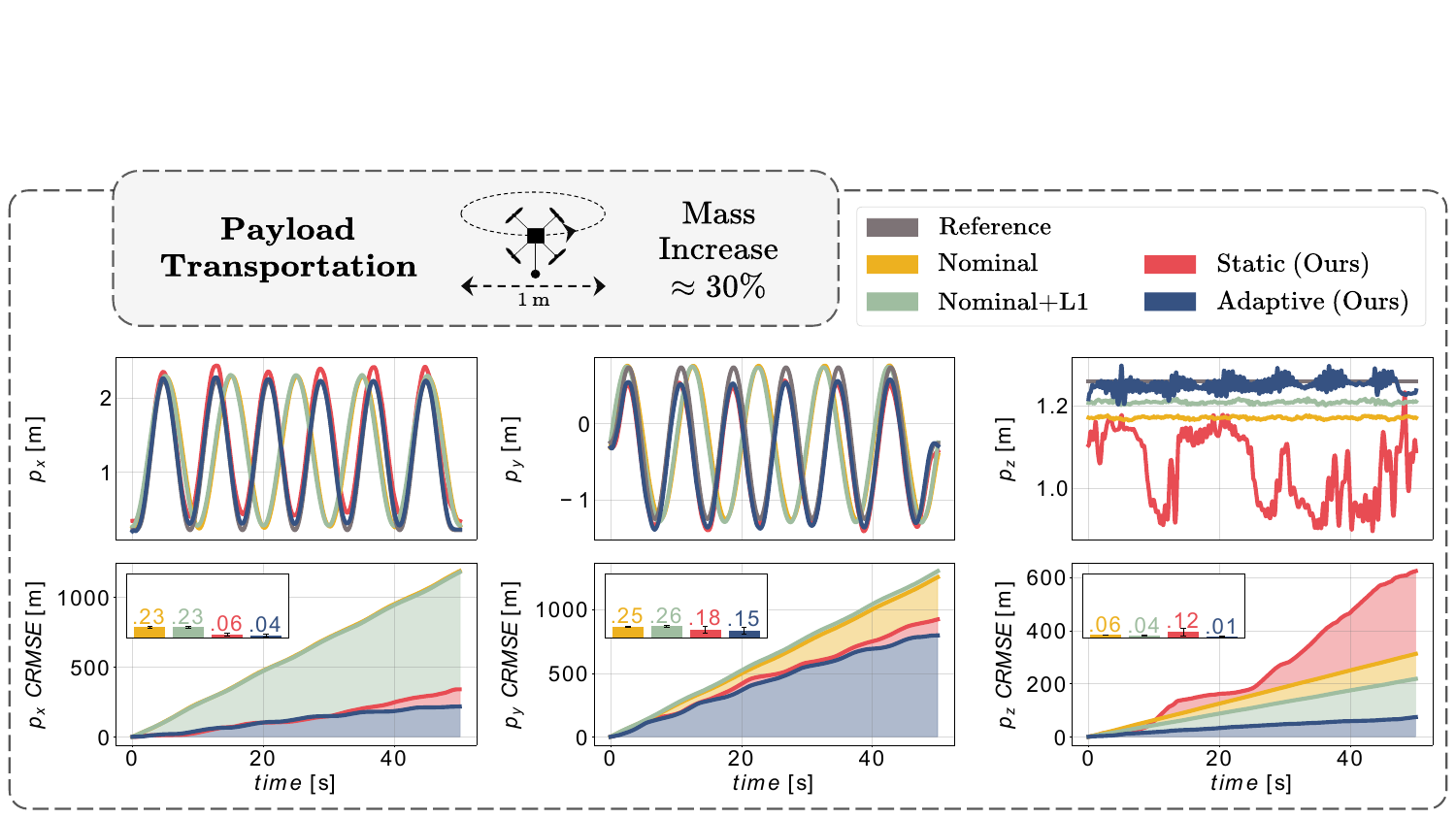}
    \includegraphics[width=0.9\linewidth, trim=0 0 0 70, clip]{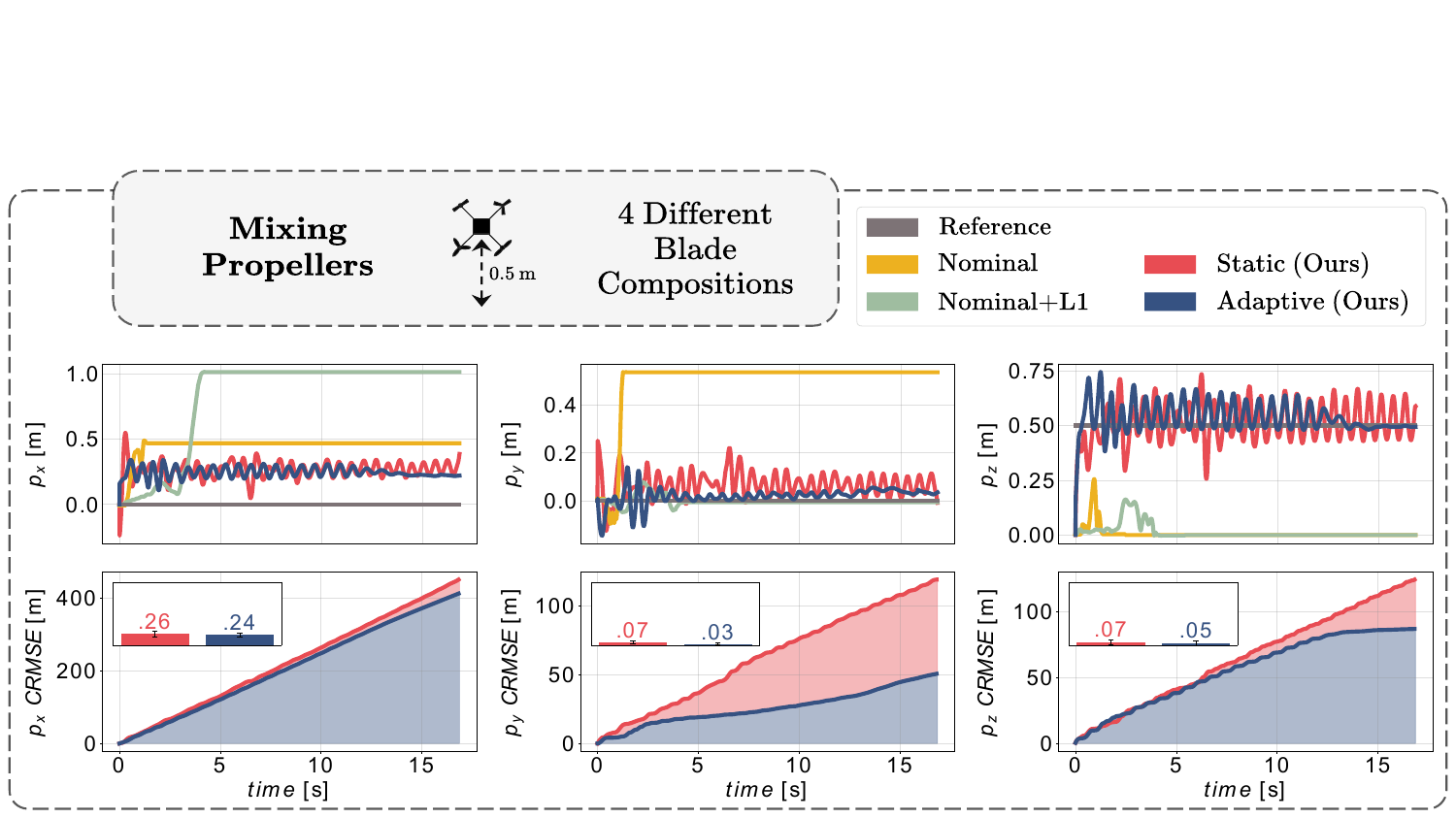}
    \includegraphics[width=0.9\linewidth, trim=0 0 0 70, clip]{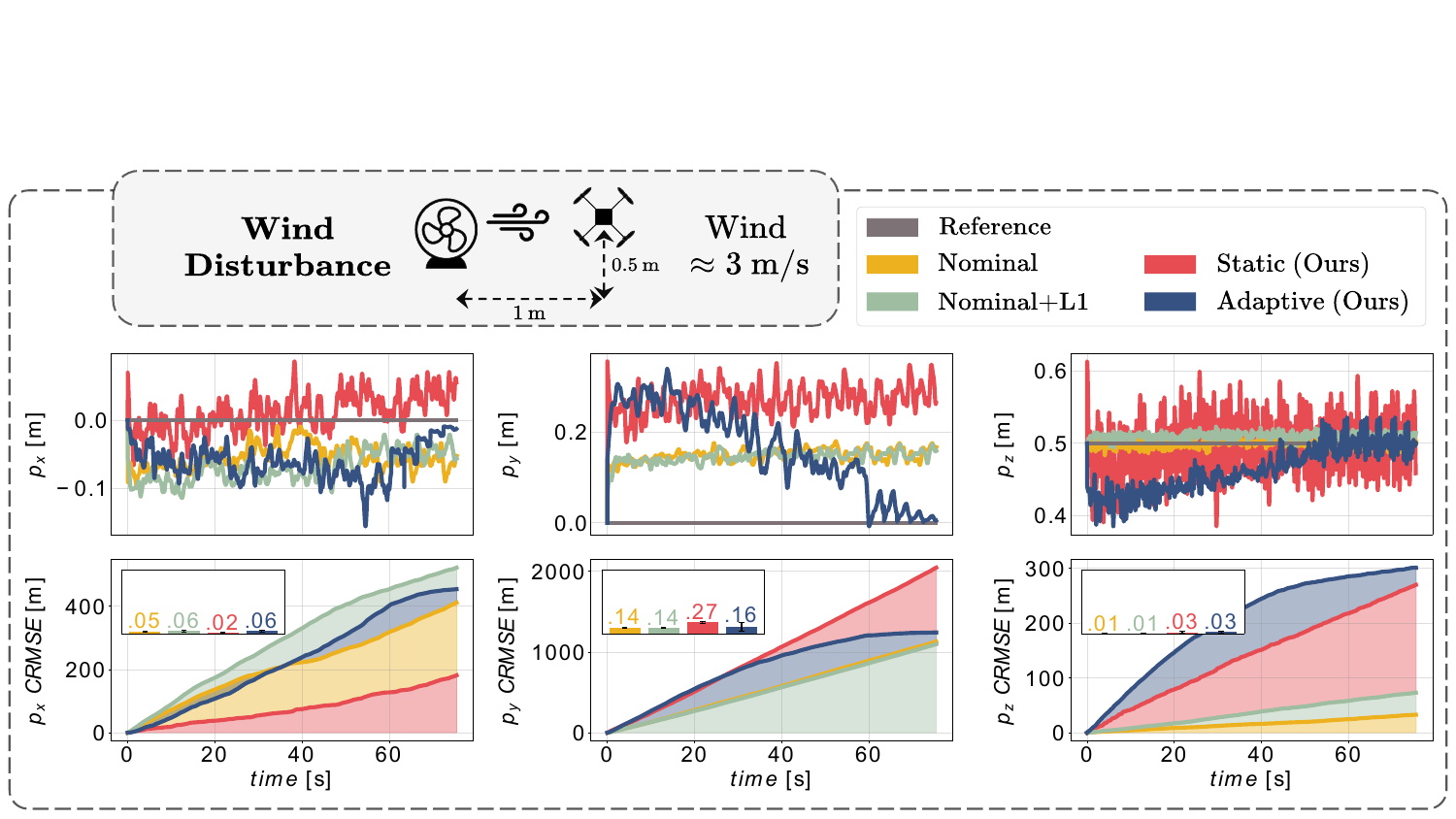}
    \caption{Role of dynamics adaptation in multiple flight operating conditions. (Inset) Average RMSE.}
    \label{fig:adapt_exp}
\end{figure*}

Figure~\ref{fig:adapt_exp} illustrates the results of these experiments.
The results demonstrate that the learned dynamics can be successfully incorporated into an uncertainty-aware MPC and combined with an online learning strategy to continuously and actively adapt to multiple challenging flight operating conditions.

When extending the system with a cable-suspended payload, non-adaptive controllers (\textit{Nominal}, \textit{Static}) struggle to accurately track the desired trajectory, which may lead to fatal navigation failures.
Contrarily, augmenting the MPC with an L1 adaptive controller allows us to continuously refine the control action to match the unmodelled disturbances and improve the tracking error by $33\%$.
However, this augmentation does not improve the dynamics model leveraged by the MPC, hence the optimization problem is still solved by respecting defective dynamics constraints.
Conversely, our approach fully leverages online learning to adapt the NN-based dynamics model to the present flight regime, and thus the MPC predictive power is fully exploited during the controller optimization, resulting in a tracking error reduced by $75\%$ with respect to the \textit{Nominal+L1} controller and $92\%$ compared to the \textit{Static} baseline.

When mixing different propellers, the quadrotor undergoes severe oscillations and may fail to track even simple near-hovering trajectories, as demonstrated by \textit{Nominal} and \textit{Nominal+L1} flight performance.
The \textit{Nominal} controller heavily relies on an old system model and fails to adapt to the new system dynamics. Consequently, after a few seconds, the quadrotor crashes.
L1 adaptive control augments the MPC predicted actions to stabilize the system, however, it is still tightly bound to the old physical dynamics and also leads to a fatal crash.
Contrarily, when leveraging NN-based models, the MPC controls the quadrotor to hover close to the desired position.
In particular, the \textit{Adaptive} controller successfully stabilizes the system, hence validating the proposed approach.
Interestingly, the learning transient during the adaptation of the NN-based dynamics does not introduce additional oscillations but leads to a stable tracking of the desired position.

Accurate tracking under challenging wind conditions is difficult due to the highly stochastic nature of wind that imposes continuous vibrations on the quadrotor, as reflected in the experimental results.
When leveraging the physics-based model, the MPC tracks the desired position approximately well.
Contrarily, the NN-based dynamics model learned offline struggles to generalize well to the new real-world dynamics
and the flight performance of the \textit{Static} baseline deteriorates compared to the \textit{Nominal} controller.
On the other hand, when employing the proposed \textit{Adaptive} controller, refining the dynamics model enables the MPC to accurately track the reference state after a minute of online learning, outperforming \textit{Nominal} even when augmented with an L1 adaptive controller.

\begin{figure*}[t]
    \centering
    \includegraphics[width=\linewidth, trim=0 10 0 60, clip]{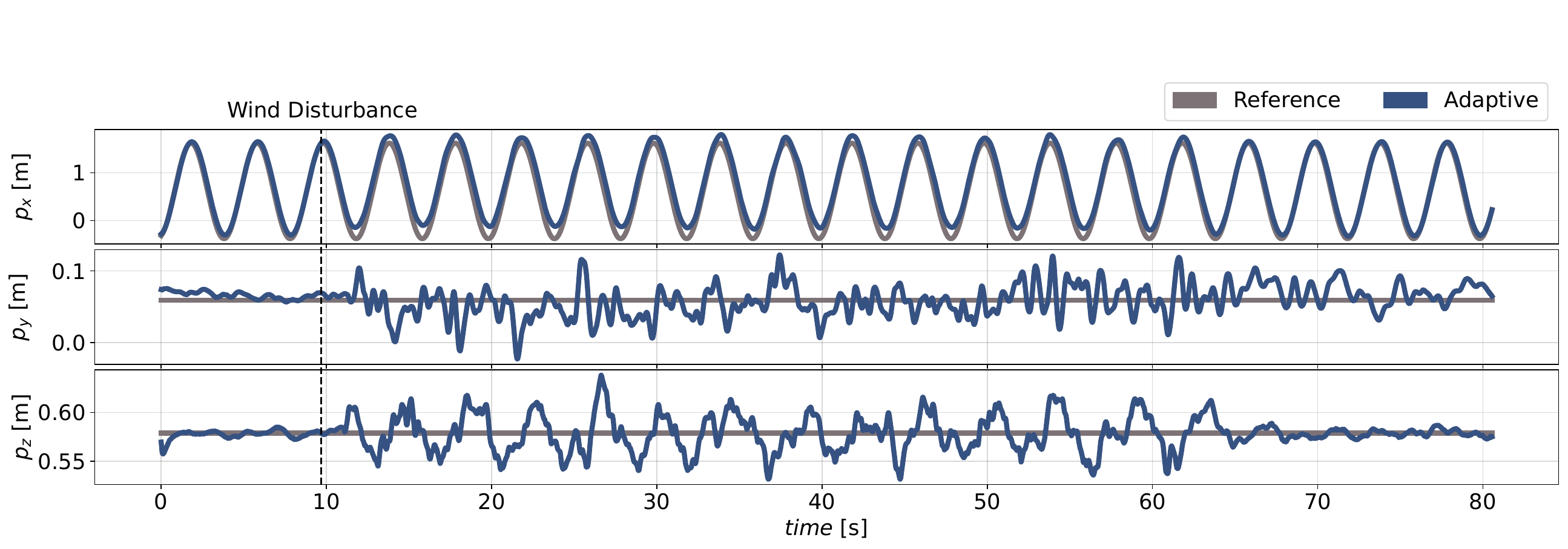}
    \vspace{-2em}
\end{figure*}

\begin{figure*}[t]
    \centering
    \vspace{-0.5em}
    \subfigure{\includegraphics[width=0.475\linewidth, trim=0 0 0 0, clip]{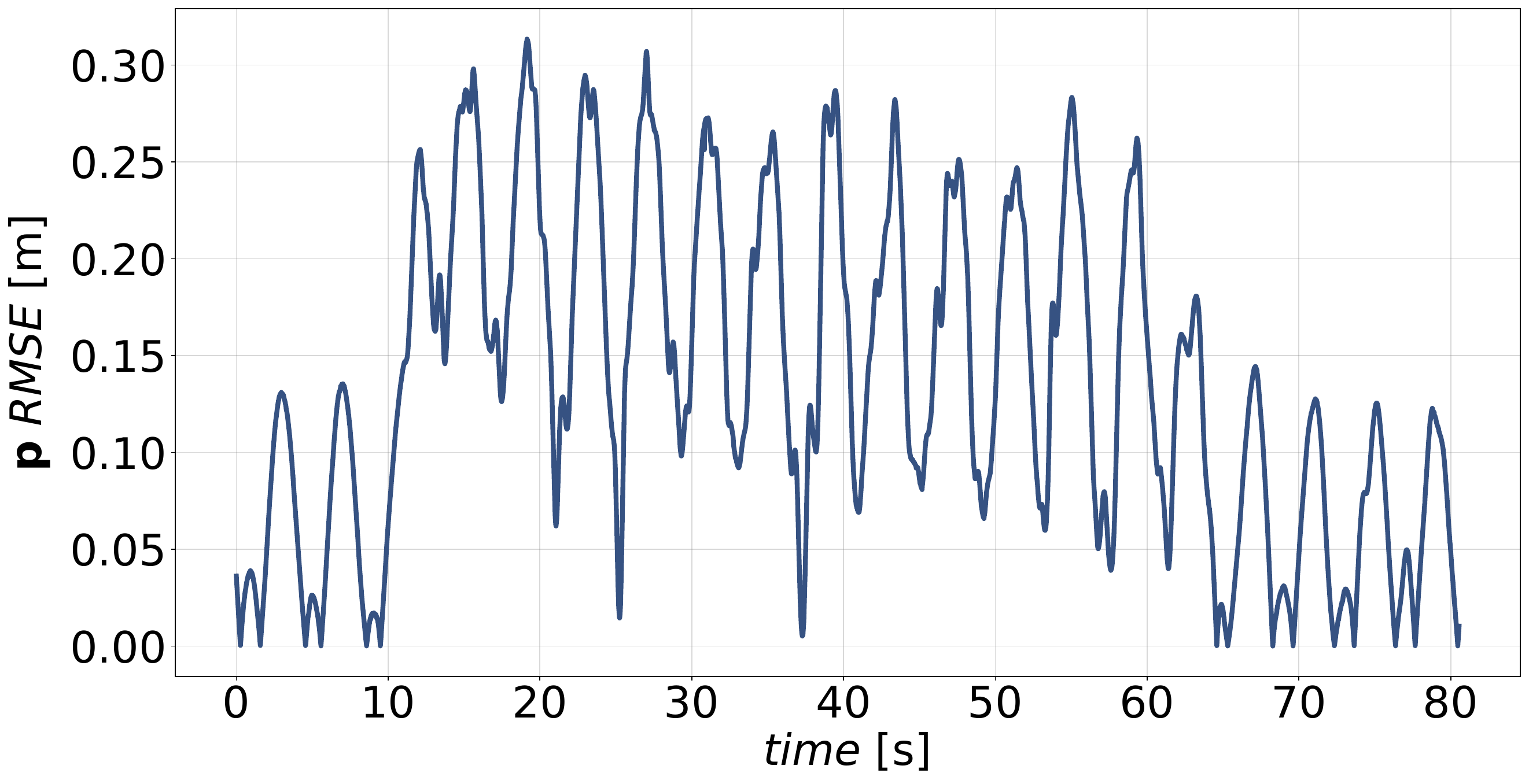}}
    \hspace{1em}
    \raisebox{1.75em}{\subfigure{\includegraphics[width=0.475\linewidth, trim=150 130 150 120, clip]{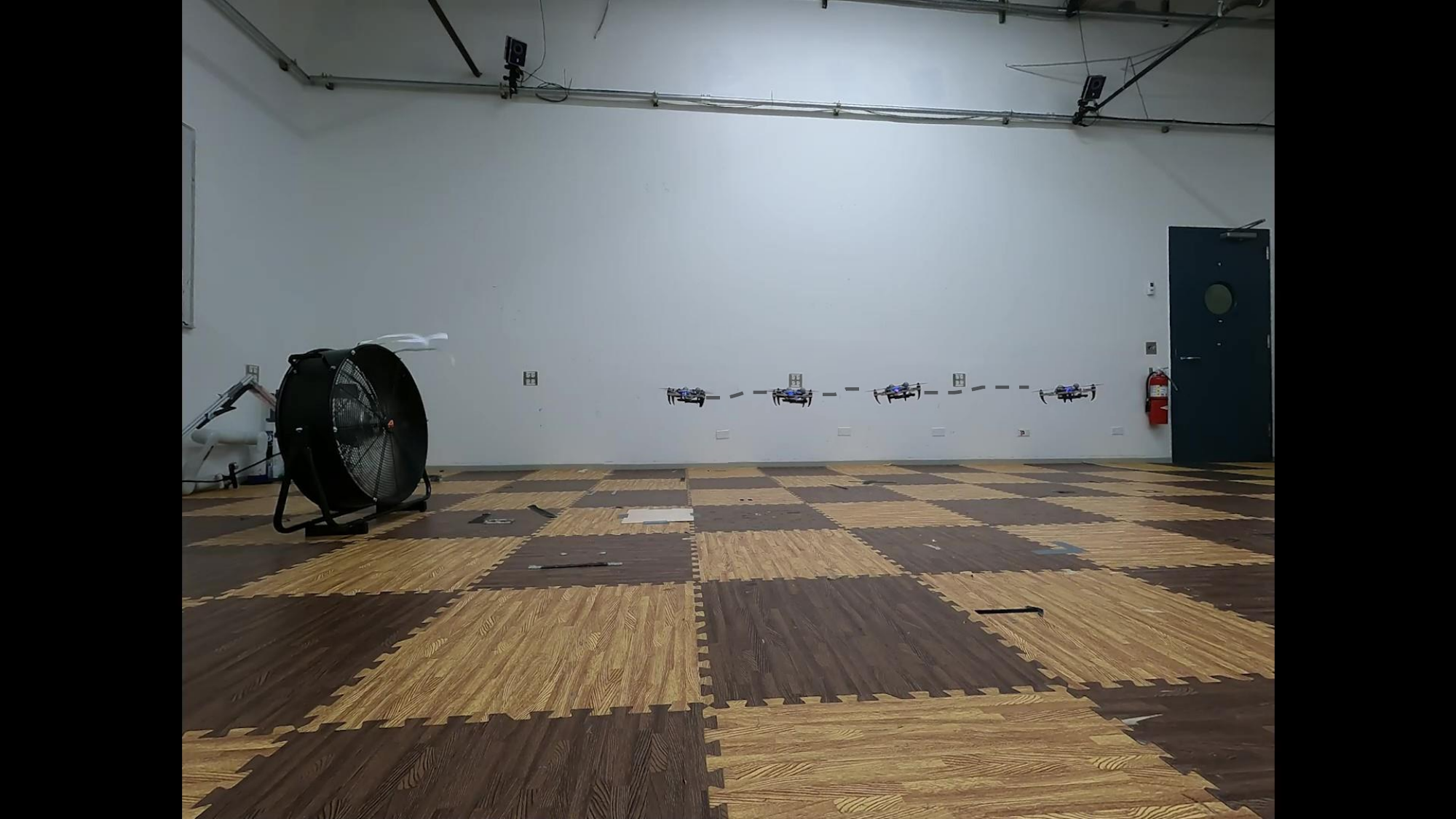}}}
    \vspace{-0.5em}
    \caption{
    \rebuttal{
    Analysis of the tracking performance convergence when an unmodeled wind disturbance is introduced into the environment. 
    Gusts of wind are directed over the $x$ axis, causing the quadrotor to overshoot and undershoot. 
    By actively adjusting its model, the quadrotor learns the new dynamics and can once again accurately track its trajectory.
    }}
    \label{fig:convergence_analysis1}
\end{figure*}

\begin{figure*}[t]
    \centering
    \includegraphics[width=0.98\linewidth, trim=0 10 0 150, clip]{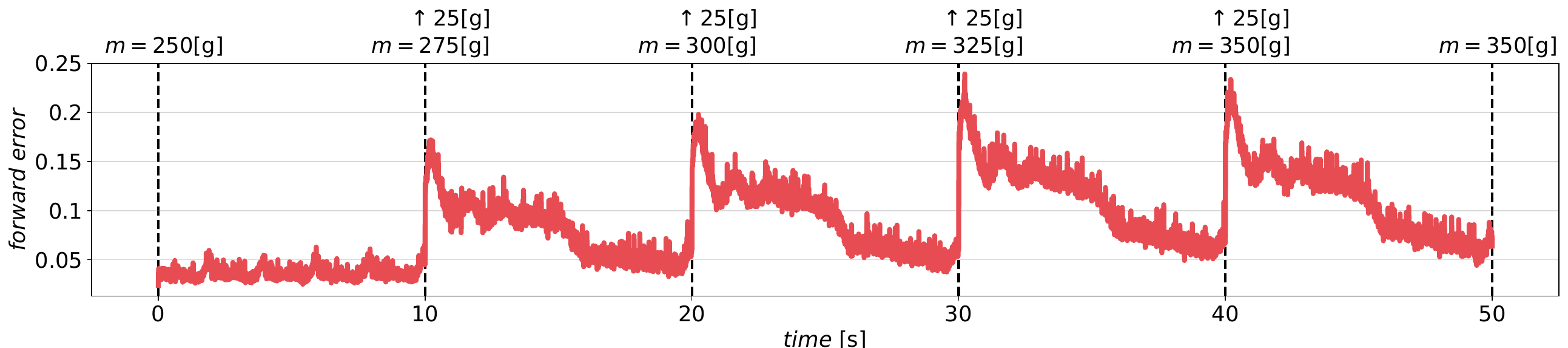}
    \vspace{-0.25em}
    \caption{\rebuttal{
    Analysis of the forward error convergence when the system's mass $m$ is increased by $25~\si{g}$ every $10~\si{s}$.
    Each time the mass is altered, the forward error increases, as the internal simulated predictions made by the MPC differ substantially from the actual state estimations.
    However, the online optimization consistently converges by minimizing the error and capturing the system configuration modification.
    This showcases the resiliency of the proposed learning approach.
    }}
    \label{fig:convergence_analysis2}
    \vspace{-0.5em}
\end{figure*}

\rebuttal{
\subsection{Convergence Analysis} \label{sec:convergence_analysis}
Theoretical guarantees of convergence are ideal in the development of learning algorithms as they ensure the reliability and predictability of the algorithm's performance under certain conditions. 
However, real-world scenarios, especially those involving quadrotors, present unique challenges, such as non-linearity, noise, disturbances, non-stationarity, and partial observability of the environment. 
These complexities often preclude the establishment of solid theoretical guarantees of convergence.
Hence, in this work, we propose an empirical approach to study the convergence properties of our approach.
We argue that the performance of an algorithm in a complex and unpredictable real-world scenario is the ultimate validation of its effectiveness.

Previous experiments have shown empirically the effectiveness of our approach in improving the model's accuracy over time, in both simulated and real-world environments. 
Building on this, we further explore the convergence capabilities of our approach in this analysis, focusing on its ability to maintain its effectiveness in the face of multiple unexpected consecutive disturbances, providing consistent tracking performance before and after it adapts to new dynamics.
Particularly, to test the algorithm's resiliency in the real world, we introduce an unmodeled wind disturbance along the quadrotor's trajectory as it follows a linear path over the $x$ axis, observing the tracking performance starting from the initial introduction of the disturbance. 
We use an uncertainty-aware MPC and a dynamics model trained on real-world data to control the quadrotor. 
Figure~\ref{fig:convergence_analysis1} shows the results of this experiment.

We continue our empirical convergence analysis by studying the ability of the learning algorithm to continuously apply its learned knowledge to new, unseen conditions consistently.
Specifically, we continually control our quadrotor to track a circular trajectory, increasing the system's weight by $25~\si{g}$ every $10~\si{s}$, and monitoring the forward error convergence over time.
We conduct this experiment in a simulation environment, using the uncertainty-aware MPC and a dynamics model trained on simulated data to control the quadrotor. 
In doing so, the dynamics of the model closely match those of the simulation environment, and any disturbances introduced in the simulator become the sole source of dynamic uncertainty for the model to adapt to. 
Figure~\ref{fig:convergence_analysis2} presents the results of this experiment.

Whenever a new disturbance appears, impacting the dynamics, the model is adjusted by optimizing the forward error. 
Upon convergence, the model aligns with the new dynamics, allowing the controller to accurately track the desired trajectory. 
This is particularly visible in the resiliency analysis, where the precise tracking performance preceding the wind disturbance is restored after the convergence of the online optimization. 
Specifically, after convergence, the overshooting and undershooting behaviors dictated by the unmodeled wind gusts cease to affect the tracking performance.

These results underscore the resilience and generalization capabilities of our proposed approach when faced with new and unseen dynamics. 
These are essential attributes for a successful online learning algorithm, as they ensure its reliable performance across a variety of conditions and scenarios. 
These results offer empirical evidence of the dynamic model's capacity for continuous improvement as well as also empirically validate the convergence of our proposed framework.
}

\begin{table*}[t]
    \centering
    \caption{\label{tab:computation_analysis}}
    \vspace{-0.75em}
    \rebuttal{
    \caption*{\scshape Computation Time Analysis~$[\SI{}{\milli\second}]$}
    \begin{tabular}{c c c c c c c c}
        \toprule\toprule
        \multirow{2}{*}{Model Architecture} & 
        \multicolumn{3}{c}{Intel Core i7-9750H} &&
        \multicolumn{3}{c}{NVIDIA Jetson Xavier NX} \\
        \cline{2-4}\cline{6-8}
        & 
        Model Prediction & 
        Model Update & 
        MPC Prediction &
        &
        Model Prediction & 
        Model Update & 
        MPC Prediction \\
        \midrule
        $ [ 64, 64, 32 ] $        & $ 0.004 $ & $ 0.001 $ & $ 284 $ &
                                  & $ 0.079 $ & $ 0.005 $ & $ 167 $ \\
        $ [ 128, 64, 32 ] $       & $ 0.007 $ & $ 0.001 $ & $ 257 $ &
                                  & $ 0.088 $ & $ 0.005 $ & $ 103 $ \\
        $ [ 128, 128, 64 ] $      & $ 0.011 $ & $ 0.001 $ & $ 185 $ &
                                  & $ 0.097 $ & $ 0.009 $ & $ 59 $ \\
        $ [ 256, 128, 128, 64 ] $ & $ 0.017 $ & $ 0.001 $ & $ 75 $ &
                                  & $ 0.121 $ & $ 0.009 $ & $ 11 $ \\
        \bottomrule\bottomrule
    \end{tabular}
    }
    \vspace{1em}
\end{table*}

\begin{table*}[t]
    \begin{minipage}{.33\linewidth}
        \setlength\tabcolsep{4.5pt}
        \centering
        \caption{\label{tab:ablating_hyperparams_1}}
        \vspace{-0.75em}
        \caption*{\scshape Tracking Perf After $1$ Loop~$[\SI{}{\meter}]$}
        \begin{tabular}{c c | c c c c}
            \toprule\toprule
            & & \multicolumn{4}{c}{Learning Rate}\\
            & & $2e-1$ & $2e-3$ & $2e-5$ & $2e-7$\\
            \midrule
            \parbox[t]{1mm}{\multirow{4}{*}{\rotatebox[origin=c]{90}{Batch Size}}}
             & $ 1$ & crash    & crash    & $ 0.44 $ & $ 0.38 $\\
             & $10$ & $ 1.22 $ & $ 0.19 $ & $ 0.10 $ & $ 0.34 $\\
             & $20$ & $ 0.95 $ & $ 0.25 $ & $ 0.11 $ & $ 0.31 $\\
             & $40$ & $ 0.52 $ & $ 0.31 $ & $ 0.18 $ & $ 0.33 $\\
            \bottomrule\bottomrule
        \end{tabular}
    \end{minipage}
    \begin{minipage}{.33\linewidth}
        \setlength\tabcolsep{4.5pt}
        \centering
        \caption{\label{tab:ablating_hyperparams_25}}
        \vspace{-0.75em}
        \caption*{\scshape Tracking Perf After $25$ Loops~$[\SI{}{\meter}]$}
        \begin{tabular}{c c | c c c c}
            \toprule\toprule
            & & \multicolumn{4}{c}{Learning Rate}\\
            & & $2e-1$ & $2e-3$ & $2e-5$ & $2e-7$\\
            \midrule
            \parbox[t]{1mm}{\multirow{4}{*}{\rotatebox[origin=c]{90}{Batch Size}}}
             & $ 1$ & crash    & crash    & $ 0.36 $ & $ 0.39 $\\
             & $10$ & crash    & $ 0.17 $ & $ 0.06 $ & $ 0.31 $\\
             & $20$ & $ 0.92 $ & $ 0.21 $ & $ 0.03 $ & $ 0.22 $\\
             & $40$ & $ 0.68 $ & $ 0.08 $ & $ 0.09 $ & $ 0.36 $\\
            \bottomrule\bottomrule
        \end{tabular}
    \end{minipage}
    \begin{minipage}{.33\linewidth}
        \setlength\tabcolsep{4.5pt}
        \centering
        \caption{\label{tab:ablating_hyperparams_50}}
        \vspace{-0.75em}
        \caption*{\scshape Tracking Perf After $50$ Loops~$[\SI{}{\meter}]$}
        \begin{tabular}{c c | c c c c}
            \toprule\toprule
            & & \multicolumn{4}{c}{Learning Rate}\\
            & & $2e-1$ & $2e-3$ & $2e-5$ & $2e-7$\\
            \midrule
            \parbox[t]{1mm}{\multirow{4}{*}{\rotatebox[origin=c]{90}{Batch Size}}}
             & $ 1$ & crash    & crash    & $ 0.27 $ & $ 0.38 $\\
             & $10$ & crash    & $ 0.09 $ & $ 0.07 $ & $ 0.31 $\\
             & $20$ & $ 1.01 $ & $ 0.16 $ & $ 0.04 $ & $ 0.21 $\\
             & $40$ & $ 0.95 $ & $ 0.12 $ & $ 0.03 $ & $ 0.37 $\\
            \bottomrule\bottomrule
        \end{tabular}
    \end{minipage}
    \vspace{1em}
\end{table*}

\rebuttal{
\subsection{Computational Complexity Analysis} \label{app:computation_analysis}
Our framework has been meticulously engineered with the objective of operating optimally on embedded platforms. 
Nonetheless, throughout the course of our study, we confronted a significant computational challenge: the implementation of the MPC on the Qualcomm\textsuperscript{\textregistered} Snapdragon\textsuperscript{\texttrademark} board. 
Despite simplifying the task to use only the \textit{Nominal} model, the hardware was incapable of conducting the required computations in real time
However, this does not discount the proficiency of the Qualcomm\textsuperscript{\textregistered} Snapdragon\textsuperscript{\texttrademark} board entirely as it is more than capable of handling the inference of our NN-based dynamics. 

To underscore the versatility and scalability of our proposed method, we expanded our testing environment. 
Progressing from laptop-grade hardware, we implemented our method on NVIDIA's Jetson Xavier NX - a popular choice for computational tasks in the field of robotics~\cite{foehn2022agilicious, saviolo2023unifying, tal2020accurate}.
The successful execution on this platform evidences that our method holds promise for scalability and adaptability to commonly utilized hardware platforms in robotics research.

To demonstrate this, we have performed a detailed analysis of the computational cost when running the entire proposed framework with different dynamics model choices on a laptop computer with an Intel Core i7-9750H CPU and on an embedded computer with an NVIDIA Jetson Xavier NX.
In particular, we have investigated the computation time required by the different dynamics models to infer a prediction and update their weights and the time required by the MPC to perform its optimization and predict an optimal action.
The results of this analysis are reported in Table~\ref{tab:computation_analysis}.
Note that the model and MPC hyper-parameters are maintained equal to the ones presented in Section~\ref{sec:exp_setup}.

The results show that the complexity of the dynamics model significantly impacts the computational time required by the MPC to solve its optimization problem.
In particular, on embedded systems, deep models are impractical for real-time inference.
This underlines the importance of tailoring the model architecture based on the specific hardware resources available. 
If the dynamics model does not ensure a minimum inference time, which would permit the MPC to compute actions at an acceptable rate (typically around $100$~$\SI{}{\hertz}$ for micro aerial vehicles), the actuation of each action commanded by the controller would be delayed. 
This delay can lead to a significant degradation in flight performance, potentially causing failures and crashes. 
Therefore, it is critical to ensure that the model complexity is suited to the hardware and control system in order to maintain the necessary real-time control performance.

Furthermore, the results highlight that the time required by the model to update its weights is negligible, even on embedded systems.
We believe that this is due to two key reasons. First, the weight update of the dynamics model entails only refining the weights and biases of the output layer. Second, the dynamics model is efficiently implemented in C++ in our custom implementation, ensuring maximum performance.

Finally, we study the relationship between the state dimension and the computational complexity of the proposed framework.
The complexity of a NN, such as the one utilized in this study, is determined by several factors, including the input size, the number of layers, the number of neurons per layer, and the type of activation function employed.
For the sake of clarity, let's consider a simplistic NN with a single hidden layer. 
The primary operations of such a model include matrix multiplication and the application of the activation function. 
Given an input vector of size $n$, a hidden layer containing $h$ neurons, and an output layer of size $o$, we can approximate the computational complexity as follows:
\begin{itemize}
    \item The input to hidden layer computation: This process involves the multiplication of an $n \times 1$ input vector with an $n \times h$ weight matrix, subsequently adding a $1 \times h$ bias vector. Therefore, the computational complexity of this operation is approximately $O(nh)$.
    \item Applying the activation function: As this function is applied individually to all $h$ neurons within the hidden layer, its computational complexity is approximately $O(h)$.
    \item The hidden to output layer computation: This process involves the multiplication of a $h \times 1$ hidden layer output vector with an $h \times o$ weight matrix, subsequently adding a $1 \times o$ bias vector. Therefore, the computational complexity of this operation is approximately $O(ho)$.
    \item Applying the activation function to the output layer: As this function is applied individually to all $o$ neurons within the output layer, its computational complexity is approximately $O(o)$.
\end{itemize}
Consequently, for a single pass or forward propagation, the overall computational complexity is roughly $O(nh + h + ho + o)$. 
As depicted, while the input size $n$ does indeed influence computational complexity, it constitutes only a fraction of the total calculation. 
The number of neurons $h$ in the hidden layer, alongside the number of output neurons $o$, considerably contributes to the overall computational complexity.
This complexity would be further multiplied by the number of samples processed in a batch, and yet again by the number of epochs for training. 
Furthermore, this complexity does not account for the backward pass used during training for backpropagation, which would approximately double the computational complexity as it involves similar computations to the forward pass.
Nevertheless, it's essential to note that while this analysis is applicable to the dynamics model, it does not apply to the computational complexity of the MPC. 
As demonstrated in previous work~\cite{saviolo2022pitcn}, the state dimension significantly impacts the controller's computation, even when optimized using ACADOS or other MPC-specific frameworks for embedded systems. 
Consequently, this places a limitation on the integration of historical information in the state, which would be highly beneficial to the learning convergence.
}

\rebuttal{
\subsection{Choosing Hyper-parameters for Online Optimization} \label{app:online_opt_ablation}
The efficacy of the learned model's online adaptation to unseen dynamics significantly depends on two hyper-parameters: batch size and learning rate. Improper setting of these parameters can hinder the model's learning ability or lead it to adapt to insignificant data noise due to faulty sensors or inaccurate state estimates. 
Therefore, we conducted a study to understand the dependency of these hyper-parameters during the model's adaptation to unseen dynamics.

In this set of experiments, we continuously control our quadrotor to track the Ellipse testing trajectory and monitor the positional RMSE after $1$, $25$, and $50$ loops. 
This experiment was conducted in a simulated environment, using the uncertainty-aware MPC and a dynamics model trained on real-world data to control the quadrotor. 
The goal of the online optimization is to adapt the dynamics model to the simulated dynamics. 
By varying batch sizes and learning rates, we could analyze their relationship with control performance. 
Our findings are presented in Tables~\ref{tab:ablating_hyperparams_1}, \ref{tab:ablating_hyperparams_25}, and \ref{tab:ablating_hyperparams_50}.

The results demonstrate a clear correlation between batch size and learning rate. 
Achieving a balance between these two parameters is crucial. 
Larger learning rates are more compatible with larger batch sizes.
Conversely, when a high learning rate is paired with small batch size, the model may undergo severe changes, which could potentially cause instability and even crashes.
In general, larger batch sizes are preferred, especially when adapting to real-world dynamics that are highly non-linear and subject to noisy state estimates. 
However, it is important to note that larger batch sizes imply slower adaptation due to the increased amount of state estimates required to compute the weight update.

Significantly, our findings show that an exceedingly low learning rate can inhibit the model from adapting appropriately, leading to suboptimal performance. 
This underscores the importance of consistently updating the dynamics model using online optimization strategies.
}

\section{Discussion}
The experimental results demonstrate that the learned dynamics can be successfully incorporated into an uncertainty-aware MPC framework and combined with an online learning strategy to continuously and actively adapt to static (e.g., changes of mass, mixing of propellers) and stochastic (e.g., payload swings, wind disturbance) model mismatches.
At the same time, the results highlight two important aspects of the proposed online optimization strategy.

First, even though the training and testing dynamics distributions are substantially different (Figure~\ref{fig:system_conf}), the learning transients do not introduce catastrophic oscillations during the online optimization.
We consider mini-batch optimization key for this behavior.
By leveraging batches of observed state-control pairs, the gradient points in the direction of smooth transitions between different operating regimes.
In general, the experimental results show a convergence of the stability of the system (i.e., the controller's initial overshooting is minimized due to a refined dynamics model). 
This behavior is well illustrated by the mixing propellers experiment and the relative clip in the supplementary video.
During the wind disturbance experiments, the stochastic nature of the wind disturbance makes the oscillating behavior difficult to minimize. However, the quadrotor is still able to accurately track the desired trajectory and significantly improve the tracking performance.

Second, during online optimization, the gradient directions strictly depend on the magnitude of the forward error on each state component.
Hence, if the error over a component is much larger than the error over the others, the update of the dynamics model will most likely improve the predictive performance over that component. 
Sometimes, this may come at the cost of marginally degrading the performance of the other components. 
However, since this degradation is minimal, this can be considered negligible. For example, during the wind disturbance experiment (Figure~\ref{fig:adapt_exp}), the forward error over the $y$ component of the position is much more significant than the error over the other state components.
Hence, the online optimization strategy focuses on minimizing the forward error over this component. 
As a result, despite the stochastic nature of wind that makes the environmental dynamics extremely difficult to model, after about $60$ seconds the quadrotor reaches the desired trajectory.
Similarly, during the payload transportation experiment, the online optimization successfully minimizes the error over the $z$ component of position due to the mass mismatch. 
However, in this experiment, the error over the $y$ component of the position is still significantly high. 
We believe that such an error is related to the stochastic oscillation of the cable-suspended payload.
A solution to this problem would be to have extra information in \rebuttal{the} input to the dynamics model about the payload position with respect to the quadrotor~\cite{belkhale2021payloadmetalearn}, hence making the state fully observable~\cite{markov1954theory}.
Consequently, the dynamics model would be able to understand if the cable is slack or tilted and find a correlation to its dynamical evolution.
However, as we do not provide this information to the dynamics model to maintain a task-agnostic approach, the optimization strategy struggles to optimize the tracking performance over the $x-y$ plane.

\section{Conclusion and Future Works}
Active learning of the system dynamics has the potential to terrifically impact the development of multiple robotic systems, enabling fast modeling and high-performance control in multiple operating conditions.
Therefore, in this work, we presented a self-supervised framework for actively learning the dynamics of nonlinear robotic systems.
\rebuttal{Our innovative approach combines MPC with aleatoric uncertainty, uniquely combining offline dynamics learning with active online adaptation.
Furthermore, we addressed the challenges related to transitioning this approach from theory to practice on SWaP-constrained robots, presenting a clear formulation that can be implemented and reproduced.}
\rebuttal{Through extensive experimentation, we demonstrated that learned dynamics continuously adapt to multiple challenging flight regimes and operating conditions, enabling unprecedented flight control.
These results provide robust empirical evidence of the proposed approach's resiliency which is critical for practical applications.}

Our results, as shown in Table~\ref{tab:prediction_error}, indicate that the proposed discrete-time dynamics model consistently surpasses a state-of-the-art continuous-time dynamics baseline~\cite{saviolo2022pitcn} in terms of predictive performance. Currently, the training process of our neural dynamics model does not take into account the impact of noisy labels~\cite{song2022survey}. Future work will delve into novel loss functions suitable for learning with noisy labels, and the influence of dataset size on the dynamics.

The ablation study presented in Figure~\ref{fig:uncertainty_ablation} validates our online learning procedure for achieving high-performance control on unseen aggressive trajectories. 
Furthermore, it shows that by conditioning the MPC with the estimated uncertainty, model learning convergence accelerates and sample efficiency improves.
Our proposed heuristic for uncertainty awareness utilizes uncertainty extracted directly from the data but does not consider uncertainty originating from limitations in the chosen dynamics model, namely epistemic uncertainty~\cite{gawlikowski2021uncertaintysurvey}. 
Future work will seek to refine the heuristic to encompass epistemic uncertainty, focusing particularly on errors caused by an insufficient model structure or lack of knowledge due to unencountered samples.

Figure~\ref{fig:adapt_exp} demonstrates that the learned dynamics can be effectively incorporated into an uncertainty-aware MPC framework and synergized with an online learning strategy for continuous adaptation to challenging operational environments. 
However, while the online learning method showed impressive performance, it demands a considerable amount of time to update the NN's weights and simultaneously maintain stable control performance. 
This is primarily due to the current strategy's reliance on a shallow NN architecture, limiting the method to updating a restricted number of weights and biases. 
Future research will investigate computationally feasible methods for utilizing deeper NN architectures that permit the updating of a larger number of parameters while keeping numerous layers frozen. 
Thus, determining the optimal number of layers to update and the sequence of updates will be a significant focus.

\rebuttal{
Figure~\ref{fig:convergence_analysis2} validates the resiliency of the proposed approach. 
At the same time, it shows a small destructive behavior due to the competing updates from the offline and online updates.
As the mass increases, the frozen backbone dynamics learned offline differ more and more from the current simulated ones.
Hybrid control is an intriguing area of future research that targets this problem, seeking to strike the optimal balance between offline and online learning phases~\cite{pinosky2022hybrid, nagabandi2018neural, chebotar2017path}.
Building on these findings, we plan to enhance our approach with hybrid learning strategies to increase learning efficiency and overall control performance.

In our work, we employed non-unit quaternions as rotation representations. While this approach is highly efficient computationally, it is not ideal for learning discrete-time dynamics due to the non-smooth manifolds it induces - it can be potentially affected by discontinuous jumps, such as quaternions flipping from +q to -q. Future work will explore other rotation representations that are smooth and minimal~\cite{peretroukhin2020smooth}.

Lastly, our extensive experiments have demonstrated that our algorithm can enhance performance over time and adapt to varying operating conditions. This provides robust empirical evidence of its generalizability and resilience, which are crucial for practical applications. Nevertheless, empirical results, while offering convincing evidence of efficacy, do not supplant the necessity for formal analysis. In the future, we aim to bolster our empirical findings with theoretical insights, thereby deepening our understanding of our algorithm's behavior and performance.
}


\section*{Appendix}

\subsection{Data Collection} \label{app:data_collection}
\rebuttal{
The data is collected by controlling the quadrotor in a series of flights both in simulation and in the real world, resulting in two datasets with analogous trajectories. 
The simulated flights are performed in the Gazebo simulator, while the real-world flights are performed in an indoor environment $10\times6\times4~\si{m^3}$ at the Agile Robotics and Perception Lab (ARPL) at the New York University. The environment is equipped with a Vicon motion capture system that allows recording accurate position and attitude measurements at $100~\si{Hz}$. 
Additionally, we record the onboard motor speeds.
Each dataset consists of $68$ trajectories with a total of $58^\prime~03^{\prime\prime}$ flight time. 
The trajectories range from straight-line accelerations to circular motions, but also parabolic maneuvers and lemniscate trajectories. All the trajectories are performed for any axis combination (i.e., $x-y$, $x-z$, $y-z$) and with different speeds and accelerations.
To capture the complex effects induced by aggressive flight, we push the quadrotor to its physical limits reaching speeds of $6~\si{\meter\per\second}$, linear accelerations of $18~\si{\meter\per\second\squared}$, \rebuttal{angular accelerations of $54~\si{\radian\per\second\squared}$,} and motor speeds of $16628~\si{rpm}$.
We recover unobserved accelerations from velocity measurements filtered by a UKF. Moreover, we filter the recorded attitudes and motor speeds measurements using a $4\textsuperscript{th}$ order Butterworth low-pass filter with a cutoff frequency of $5$.
Finally, we randomly select $62$ trajectories for training, while using the remaining $6$ for testing (Figure~\ref{fig:test_trajs}).
}

\subsection{Discrete vs. Continuous-Time Dynamics} \label{app:pred_perf}
\begin{figure}[t]
    \centering
    \includegraphics[width=\linewidth, trim=0 0 50 0, clip]
    {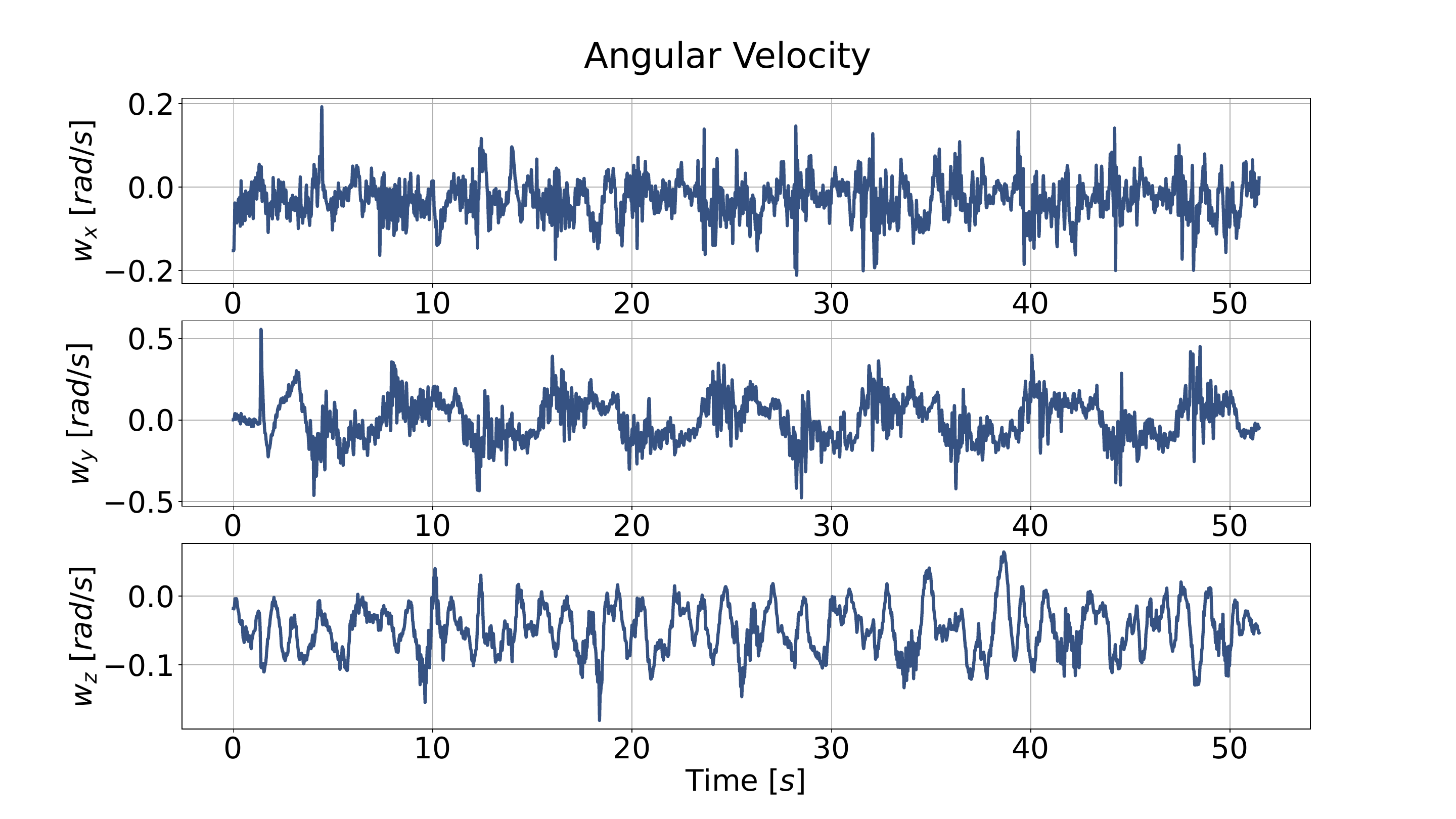}
    \includegraphics[width=\linewidth, trim=0 0 50 0, clip]{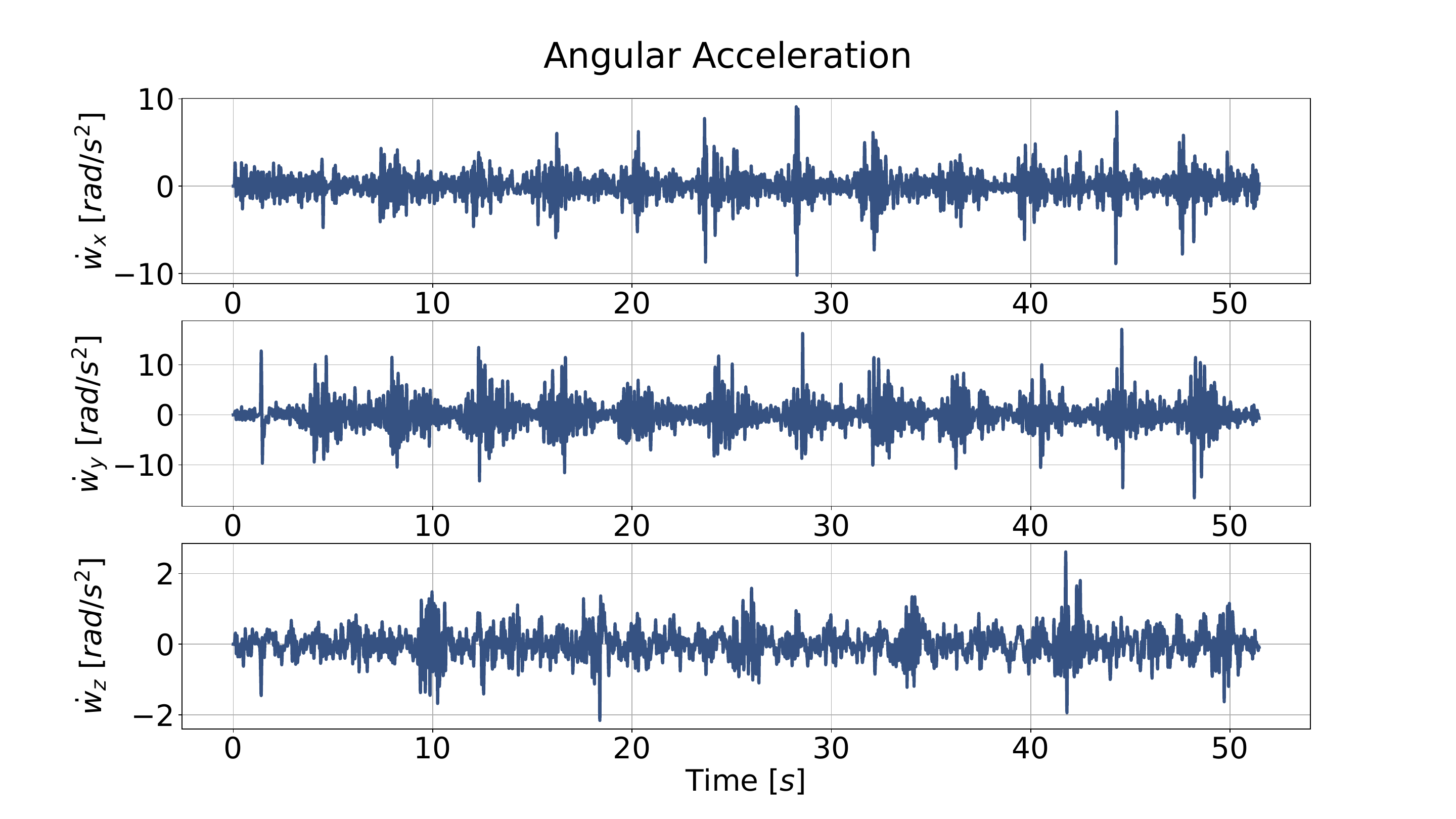}
    \caption{Samples of collected rotational velocities and accelerations from~\cite{saviolo2022pitcn} data set.}
    \label{app:data_collection_our}
\end{figure}

\begin{figure}[t]
    \centering
    \includegraphics[width=\linewidth, trim=0 0 50 0, clip]{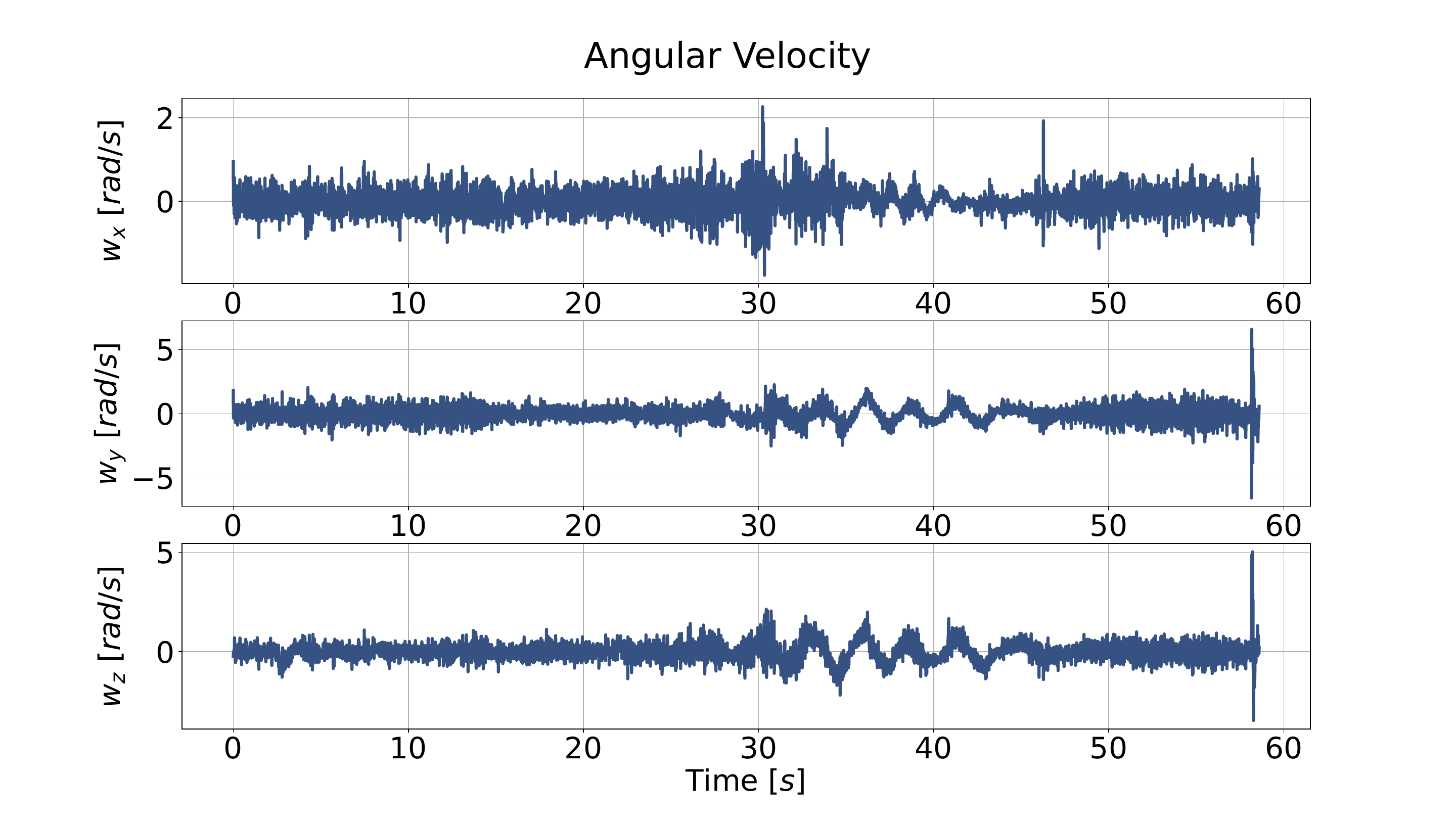}
    \includegraphics[width=\linewidth, trim=0 0 50 0, clip]{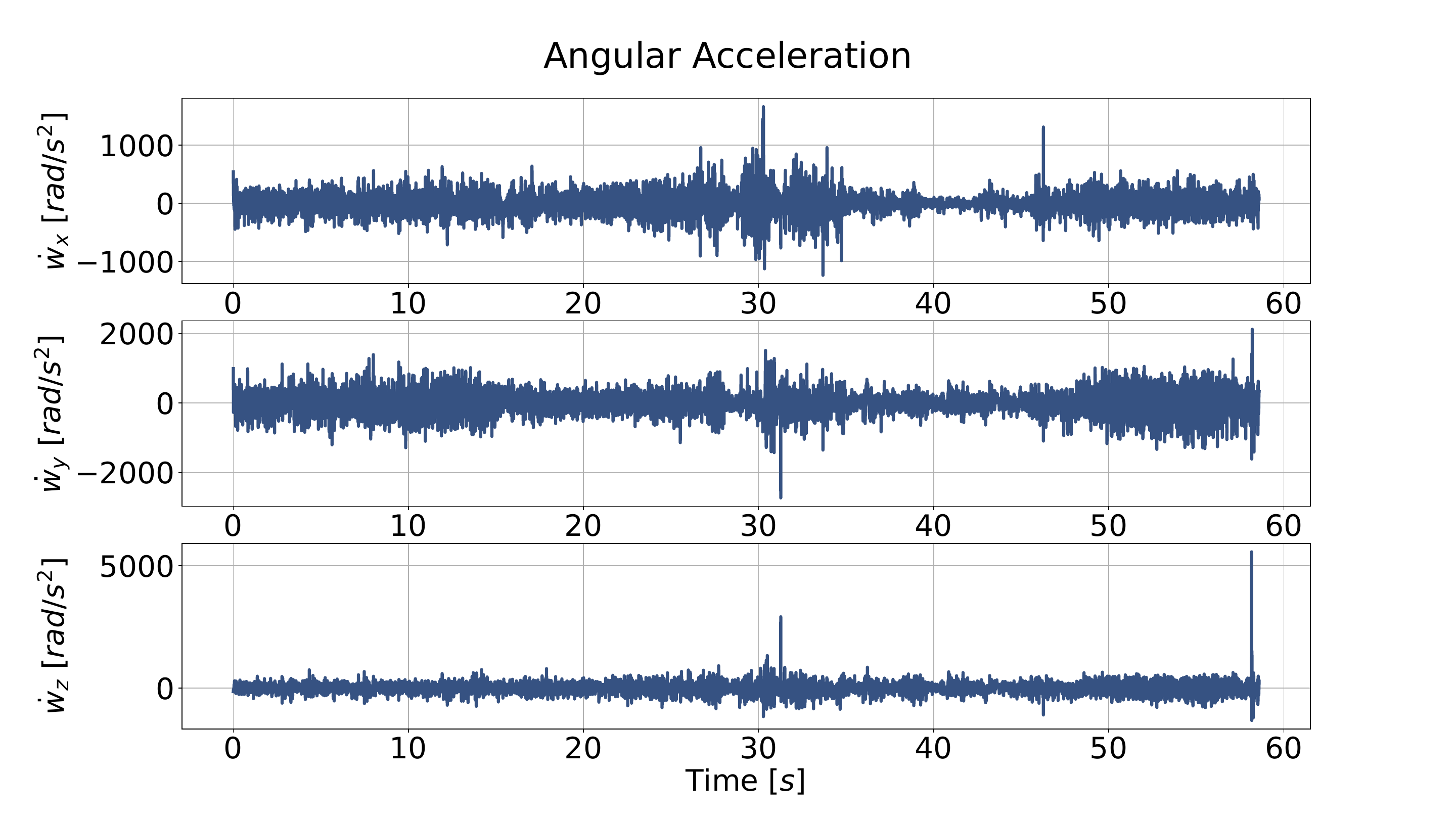}
    \caption{Samples of collected rotational velocities and accelerations from~\cite{bauersfeld2021neurobem} data set.}
    \label{app:data_collection_neurobem}
\end{figure}

We extend the analysis of discrete and continuous-time dynamics introduced in Section~\ref{sec:pred_perf}.
Specifically, we illustrate the different ground truth data that the NN models are trained on in the two different tasks.
Figure~\ref{app:data_collection_our} and Figure~\ref{app:data_collection_neurobem} illustrate samples of rotational velocities and accelerations that constitute our and \cite{bauersfeld2021neurobem} data sets. 
The rotational velocities are extracted from the IMU sensor and are thus affected by very little noise. 
Therefore, learning discrete\rebuttal{-time} dynamics is better suited for the model learning process.
In the continuous-time case, the unobserved rotational accelerations are calculated as the first-order derivative of the rotational velocities.
Hence, the mild noise affecting the velocities is strongly inflated by the differentiation operation.
Despite the low-pass filters that can be employed to reduce the noise, directly training on such noisy labels significantly deteriorates the learning process, leading to a less stable training convergence.
This is also demonstrated by the experimental results reported in Section~\ref{sec:pred_perf}.

\rebuttal{
\subsection{Dynamics Discretization} \label{app:discretizing_dynamics}
Robotic systems, such as quadrotors, operate in continuous time, which ideally requires a continuous dynamical model. However, in practice, a continuous model adds excessive complexity to robot control, making discretization of the dynamics necessary for numerical computation~\cite{bauersfeld2021neurobem, saviolo2022pitcn}.

There are several methods that can be used to discretize continuous-time dynamics of quadrotors, including Euler~\cite{hahn1991modified}, Runge-Kutta~\cite{cartwright1992dynamics}, and Trapezoidal~\cite{hammer1955trapezoidal}.
However, even the most advanced methods may introduce precision errors and result computationally expensive.

In this work, we conducted experiments to compare the learning performance of a NN-based dynamical model when extrapolating continuous-time or discrete-time dynamics from collected data.
It is important to note that even in the case of continuous-time dynamics, the model is still discretized before being used by the controller, which involves an additional discretization step.
However, by directly learning discrete-time dynamics, the discretization step is bypassed, and the NN-based dynamics can infer the discrete state at the next time step directly. This results in higher predictive performance and better control, which has been demonstrated in our experiments.
}

\subsection{Beyond The Sim-to-Real Gap} \label{app:sim2real_trap}
Simulators are valuable tools for testing quadrotors in a safe and efficient manner. 
Modern simulators can replicate complex effects like drag forces, communication delays, and random disturbances. 
However, it is practically impossible to capture the full range of real-world variability. 
Quadrotor dynamics have highly nonlinear dynamics that include stochastic effects like aerodynamic forces, torques, propeller interactions, vibrations, and structural flexibility. 
Modeling the full range of these effects accurately is extremely challenging. 
Moreover, environmental factors such as wind, humidity, and temperature directly impact quadrotor flight dynamics and stability. 
Finally, sensor measurements from quadrotors can be affected by noise and other sources of variability, which can impact the accuracy of state estimation and control algorithms. 
While simulators are valuable, they cannot fully replicate all the dynamic and complex factors that can affect the real-world performance of quadrotors. 
Therefore, accurately capturing the real-world dynamics of the quadrotor is essential for modeling their system accurately.

\bibliographystyle{IEEEtran}
\bibliography{bibliography}

\begin{IEEEbiography}[{\includegraphics[width=1in,height=1.25in,clip,keepaspectratio]{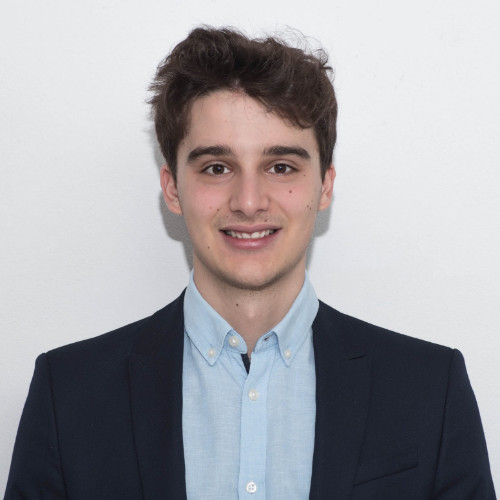}}]{Alessandro Saviolo} is a Ph.D. candidate at the Agile Robotics and Perception Lab at New York University, focusing on the advancement of AI-driven robotics, particularly in the realm of perpetual, resilient autonomy. He earned his Master's in Computer Engineering from the University of Padova, and his graduate research spanned across National Chao Tung University in Taiwan and the University of Zurich in Switzerland. Following his Master's degree, he worked as a software engineer at Flexsight, a start-up specializing in autonomous perception solutions.\end{IEEEbiography}

\begin{IEEEbiography}[{\includegraphics[width=1in,height=1.25in,clip,keepaspectratio]{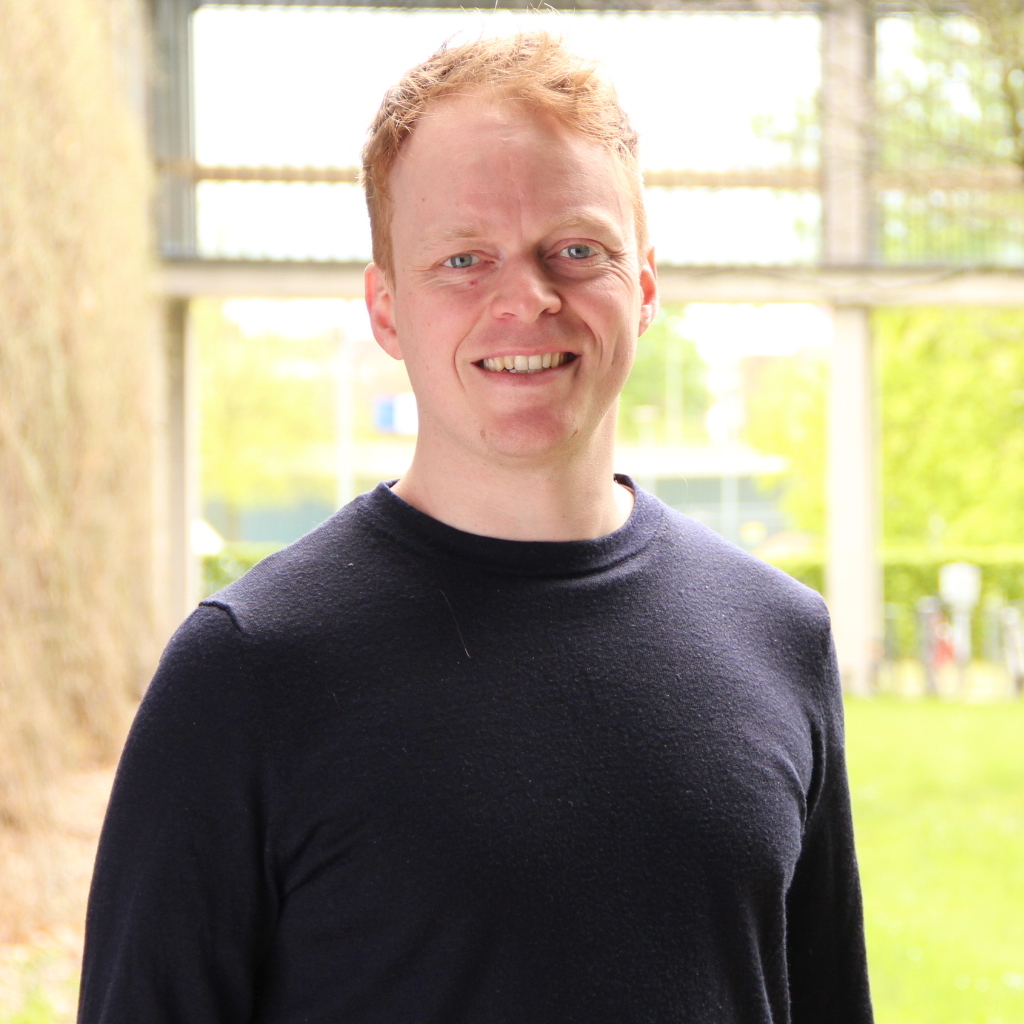}}]{Jonathan Frey} studied Mathematics at the Technical University Ilmenau and at University Freiburg, where he obtained his M.Sc. in 2019. He is currently a PhD student at University Freiburg working on numerical algorithms and advantageous problem formulations for fast MPC applications under the supervision of Prof. Dr. Moritz Diehl. He is currently the maintainer of the open-source software framework “acados” which implements fast and embedded solvers for nonlinear optimal control.\end{IEEEbiography}

\begin{IEEEbiography}[{\includegraphics[width=1in,height=1.25in,clip,keepaspectratio]{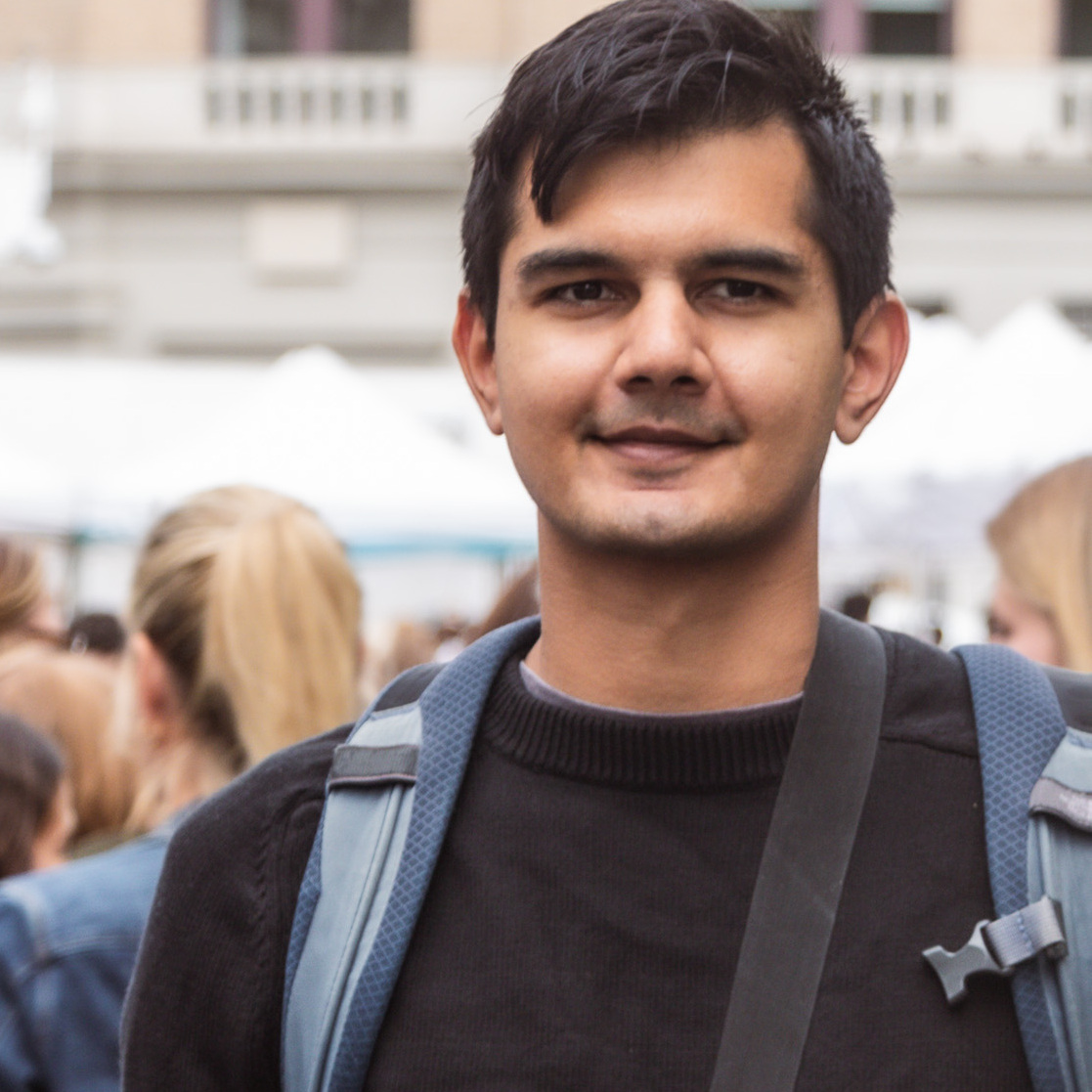}}]{Abhishek Rathod} is a Robotics Engineer at the National Robotics Engineering Center, working in the areas of control and planning for autonomous aerial systems. He earned his Master's in Robotics from New York University in 2022 and Bachelor's degree in Mechanical Engineering from the University of Idaho in 2020. 
During and following his Master's degree, he worked as a researcher at the Agile Robotics and Perception Lab, focusing on data-driven model-based control and safety-critical control with control barrier functions.
\end{IEEEbiography}

\begin{IEEEbiography}[{\includegraphics[width=1in,height=1.25in,clip,keepaspectratio]{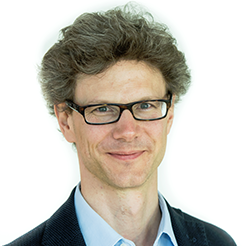}}]{Moritz Diehl} graduated with a Diploma in physics and mathematics at Heidelberg University, Heidelberg, Germany and Cambridge University, Cambridge, U.K. in 1999, and received his Ph.D. degree from Heidelberg University in 2001. From 2006 to 2013, he was a Professor with the Department of Electrical Engineering, KU Leuven, Leuven, Belgium. In 2013, he moved to the University of Freiburg, Freiburg, Germany, where he is currently the Head of the Systems Control and Optimization Laboratory, Department of Microsystems Engineering, and the Department of Mathematics.\end{IEEEbiography}

\begin{IEEEbiography}[{\includegraphics[width=1in,height=1.25in,clip,keepaspectratio]{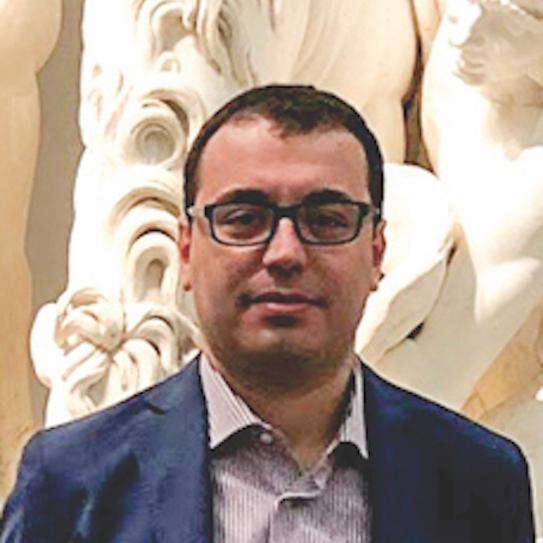}}]{Giuseppe Loianno} received the Ph.D. degree in robotics from the University of Naples "Federico II," Naples, Italy, in 2014. He is currently an Assistant Professor with New York University, New York, NY, USA, and the Director of the Agile Robotics and Perception Lab working on autonomous robots. Prior to joining NYU, he was a Postdoctoral Researcher, Research Scientist, and Team Leader with GRASP Lab, University of Pennsylvania, Philadelphia, PA, USA. He has authored or coauthored more than 70 conference papers, journal papers, and book chapters. His research interests include perception, learning, and control for autonomous robots. Dr. Loianno was the recipient of the NSF CAREER Award in 2022, DARPA Young Faculty Award in 2022, IROS Toshio Fukuda Young Professional Award in 2022, Conference Editorial Board Best Associate Editor Award at ICRA 2022, and Best Reviewer Award at ICRA 2016, and he was selected as Rising Star in AI from KAUST in 2023. He is also the Co-Chair of the IEEE RAS Technical Committee on Aerial Robotics and Unmanned Aerial Vehicles. He was the General Chair of the IEEE International Symposium on Safety, Security and Rescue Robotics (SSRR) in 2021 as well as Program Chair in 2019, 2020, and 2022. His work has been featured in a large number of renowned international news and magazines.\end{IEEEbiography}

\end{document}